\documentclass[10pt,twocolumn,letterpaper,pagebackref,breaklinks,colorlinks,allcolors=wacvblue]{article}

\usepackage{style/wacv}              

\definecolor{wacvblue}{rgb}{0.21,0.49,0.74}
\usepackage[pagebackref,breaklinks,colorlinks,allcolors=wacvblue]{hyperref}

\usepackage{microtype}
\usepackage{enumitem}
\usepackage{graphicx}
\usepackage{booktabs} 
\usepackage[table]{xcolor} 
\usepackage{pgfplots}
\pgfplotsset{compat=1.18}
\usepackage{pgfplotstable} 
\usepackage{tikz}
\usepgfplotslibrary{groupplots}
\usepackage{amssymb} 
\definecolor{mycolor1}{RGB}{220,230,241}
\definecolor{mycolor2}{RGB}{200,220,240} 
\definecolor{mycolor3}{RGB}{180,210,235} 
\newcommand{\gcell}[1]{\cellcolor{gray!12}{#1}}
\definecolor{lightgray}{gray}{0.6}
\usepackage{multirow} 
\usepackage{algorithm}
\usepackage{algorithmic}
\usepackage{pifont}     
\newcommand{\cmark}{\ding{51}} 
\newcommand{\xmark}{\ding{55}} 

\definecolor{algcomment}{RGB}{110,110,110}

\usepackage{xspace}

\usepackage{threeparttable}
\usepackage{makecell}
\usepackage{tabularx}
\usepackage{colortbl}
\usepackage{xcolor}

\usepackage{amsmath}
\usepackage{amssymb}
\usepackage{mathtools}
\usepackage{amsthm}

\usepackage[capitalize,noabbrev]{cleveref}

\theoremstyle{plain}

\theoremstyle{definition}

\theoremstyle{remark}

\usepackage{xspace}

\newcommand{\boldpar}[1]{\smallskip\noindent\textbf{#1.}\xspace}

\definecolor{nord_red}{HTML}{bf616a}
\definecolor{nord_green}{HTML}{7c906b}
\definecolor{nord_blue}{HTML}{5e81ac}
\definecolor{nord_gray}{HTML}{fafafa}
\definecolor{nord_orange}{HTML}{d08770}
\definecolor{nord_violet}{HTML}{b48ead}

\def\wacvPaperID{} 
\def\confName{}
\def\confYear{}

\title{WISE-ATTA: When to Ask for Labels in Budgeted Active Test-Time Adaptation}

\author{
Muhammad Huzaifa \quad Lea Sch\"onherr \quad Thorsten Eisenhofer\\
CISPA Helmholtz Center for Information Security\\
{\tt\small \{muhammad.huzaifa, schoenherr, eisenhofer\}@cispa.de}\\
{\small Code: \url{https://github.com/Muhammad-Huzaifaa/WISE-ATTA}}
}

\begin{document}
\maketitle

\begin{abstract}
Active test-time adaptation (ATTA) improves robustness under distribution shift by updating a deployed model during inference while selectively querying supervision. However, most existing ATTA methods implicitly assume that supervision can be requested for every incoming test batch, which can incur substantial annotation cost over long test streams.
In this work, we introduce \emph{budgeted ATTA} in which labels are available for only a fraction of test batches. This formulation shifts the central challenge from deciding \emph{what} to label within a batch to deciding \emph{when} supervision should be applied over time.
To address this challenge, we propose a budget-aware approach \emph{WISE-ATTA} that allocates supervision over the test stream based on lightweight signals computed online, prioritizing periods where supervision is likely to be most useful. When a batch is selected for supervision, we further employ a drift-based sample selection criterion that targets samples exhibiting ongoing, unconverged adaptation dynamics, enabling effective updates from a single labeled example.
We evaluate this approach on synthetic corruptions (ImageNet-C) and natural distribution shifts (ImageNet-R/K/A). Across settings, WISE-ATTA achieves competitive or improved performance compared to recent ATTA methods while requiring substantially fewer labels.
Overall, we find that the timing of supervision is a key, yet underexplored, aspect of active test-time adaptation.
\end{abstract}

\section{Introduction}
\label{intro}

Modern machine learning models are routinely deployed in environments where the test distribution differs from training, for example, due to noise, changes in data acquisition, or broader domain shifts~\cite{hendrycks2019benchmarking,recht2019imagenet,ovadia2019can}.
Test-time adaptation (TTA) addresses this challenge by updating a pretrained model on the unlabeled test stream as it becomes available, often using per-batch objectives based on entropy minimization or pseudo-label-based self-training~\cite{wang2020tent,wang2022continual}.
While effective over short horizons, purely unsupervised TTA can become unstable over long streams: small adaptation errors can accumulate, internal representations may drift, and performance can deteriorate as the distribution evolve~\cite{wang2022continual,niu2023towards}.

\begin{table}[t]
\centering
\caption{Comparison of ATTA methods on ImageNet-K. WISE-ATTA achieves the lowest average error with fewer labels.}
\label{tab:method_comparison}

\footnotesize
\renewcommand{\arraystretch}{1.12}
\setlength{\tabcolsep}{5pt}

\begin{tabularx}{\columnwidth}{
    @{}
    l
    >{\centering\arraybackslash}X
    >{\centering\arraybackslash}X
    >{\centering\arraybackslash}X
    >{\centering\arraybackslash}X
    @{}
}
\toprule
\textbf{Method}
& \shortstack{\textbf{Replay}\\\textbf{Buffer}}
& \shortstack{\textbf{Labels}\\\textbf{/ Batch}}
& \shortstack{\textbf{Batch}\\\textbf{Sel.}}
& \shortstack{\textbf{Avg.}\\\textbf{Err.} $\downarrow$} \\
\midrule

CEMA~\cite{chen2024towards}
& \cmark & teacher & \xmark & 59.2 \\

SimATTA~\cite{gui2024active}
& \cmark & 3 & \xmark & 58.2 \\

HILTTA~\cite{li2024exploring}
& \xmark & 3 & \xmark & 58.1 \\

EATTA~\cite{wang2025effortless}
& \xmark & 1 & \xmark & 58.0 \\

\midrule

\rowcolor{gray!12}
\textbf{WISE-ATTA (Ours)}
& \textbf{\xmark}
& $\mathbf{\leq 0.5}$
& \textbf{\cmark}
& \textbf{57.2} \\

\bottomrule
\end{tabularx}
\end{table}

In response, active test-time adaptation (ATTA) augments TTA with sparse supervision, providing corrective signals that anchor the adaptation process and reduce error accumulation~\cite{gui2024active,li2024exploring,wang2025effortless}.
In practice, however, test-time supervision is costly: labels may require human experts or expensive teacher models~\cite{li2024exploring,chen2024towards}, and even low annotation rates can accumulate substantial cost over long deployments~\cite{wang2025effortless}. Moreover, not all test batches contribute equally to adaptation in a continual stream. This raises a central question: \emph{how should limited supervision be used during a continual test stream?}

Thus far, existing ATTA methods implicitly adopt a batch-centric view of supervision. They assume that every incoming test batch is annotated and focus on reducing the number of labeled samples \emph{within} each batch, for example by selecting representative~\cite{gui2024active} or highly learnable samples~\cite{wang2025effortless}.
While effective at improving label efficiency locally, this formulation ties supervision rigidly to the batch structure of the test stream and applies it uniformly over time. Under a fixed global budget, such uniform allocation can be inefficient, as the benefit of supervision can vary substantially across batches and over the course of adaptation. As a result, existing methods leave open the question of \emph{when} supervision should be applied in a continual test stream.

In this paper, we address this question and consider a \emph{budgeted} ATTA setting in which supervision is available only for a limited subset of batches over a long test stream. This setting better reflects practical settings, where annotation incurs non-negligible cost and supervision may be available only intermittently due to human, computational, or latency constraints. Importantly, it changes the nature of active adaptation.
Rather than only deciding \emph{what} to label within a batch, effective adaptation now requires making online decisions about \emph{when} a batch should be labeled. We show that explicitly reasoning about the timing of supervision plays a central role in effectively using a limited budget.

To this end, we introduce \emph{WISE-ATTA}, a budget-aware approach to active test-time adaptation that estimates the utility of incoming batches and allocates a global label budget over time. As shown in \cref{tab:method_comparison}, WISE-ATTA requires neither a replay buffer nor a teacher model, and can achieve stronger performance with fewer labels than prior ATTA methods.

A key challenge in this setting is to assess the potential benefit of supervision at the moment a batch becomes available. Such decisions must be made online and without access to ground-truth labels, relying only on statistics computed from the current model and the incoming data.
WISE-ATTA addresses this with a two-stage strategy. First, it allocates supervision across the test stream by identifying batches where supervision is likely to reinforce ongoing, stable adaptation. Second, when a batch is selected, it chooses a single informative sample by comparing the model's current predictions to a temporally smoothed exponential moving average anchor. Samples whose predictions continue to drift relative to this anchor indicate that adaptation has not yet stabilized, making them promising candidates for supervision.

We evaluate WISE-ATTA on both synthetic corruptions and natural distribution shifts, including ImageNet-C/R/K/A. Across these settings, WISE-ATTA matches or improves upon recent ATTA baselines while using substantially fewer labels, demonstrating that explicitly accounting for when supervision is applied can improve label efficiency under limited annotation budgets. 

In summary, we make the following contributions:
\begin{itemize}[leftmargin=*, itemsep=0em, topsep=0em]
    \item \textbf{Budgeted active test-time adaptation.} We formulate a practical \emph{budgeted} ATTA setting where supervision is available only for a subset of test batches, shifting the focus from \emph{what} to label to \emph{when} to label.
    \item \textbf{Selective supervision under label budgets.} We propose a budget-aware approach that selectively allocates supervision over long test streams and identifies informative samples for supervision.
    \item \textbf{Evaluation of efficacy.} We demonstrate competitive performance compared to state-of-the-art ATTA methods on ImageNet-C/R/K/A with up to 50\% fewer labels.
\end{itemize}
\section{Test-Time Adaptation}
\label{sec:prelim}

Machine learning models deployed in the real world often face distribution shifts that cannot be fully anticipated at training time~\cite{hendrycks2019benchmarking,recht2019imagenet,ovadia2019can}. There are various ways to address this, such as improving robustness during training~\cite{madry2017towards,hendrycks2019augmix,sagawa2019distributionally} or continually retraining models as new data becomes available~\cite{parisi2019continual,de2021continual,kirkpatrick2017overcoming,rebuffi2017icarl,lopez2017gradient,rolnick2019experience}. However, these approaches can be impractical when distributions evolve continuously, retraining is costly, or supervision is limited at deployment.
Test-time adaptation (TTA) offers an alternative by allowing models to adapt directly during deployment using the incoming stream of test data~\cite{wang2020tent,sun2020test,wang2022continual,liang2024comprehensive}. 
In its standard form, it operates without access to ground-truth labels and performs lightweight online updates as test samples arrive.

\subsection{Online Test-Time Adaptation}
\label{sec:prelim-setup}
\label{sec:prelim-tta}

We consider TTA under a streaming evaluation protocol. A model is pretrained on a source domain and then deployed in a target environment, where test samples arrive sequentially in batches.
Let $\{\mathcal{B}_t\}_{t=1}^{T}$ denote the stream of test batches, where each batch $\mathcal{B}_t = \{x_t^i\}_{i=1}^{n}$ contains $n$ unlabeled inputs observed at time $t$, and where $T$ is typically unknown.
We assume a non-stationary environment in which the test distribution drifts over time, motivating continual adaptation.
The model consists of a feature extractor $f_{\theta_f}$ with parameters $\theta_f$ and a classifier $h_{\theta_c}$ with parameters~$\theta_c$, producing logits and class probabilities
\begin{equation*}
    z_t^i = h_{\theta_c}\!\big(f_{\theta_f}(x_t^i)\big)\,,
\end{equation*}
where $C$ is the number of classes. After observing each batch, the learner updates (a subset) of model parameters before proceeding to the next batch.

\boldpar{Entropy minimization}
The standard objective for this update step is entropy minimization~\cite{wang2020tent}. 
It is motivated by the observation that many test-time shifts primarily reduce prediction confidence without changing the predicted class.
Encouraging sharper predictions allows the model to adapt to mildly shifted samples without requiring labels.
For a given batch $\mathcal{B}_t$, the entropy objective is defined as 
\begin{equation*}
    L_{\mathrm{ent}}^t
    = \frac{1}{n}\sum_{i=1}^{n} H(p_t^i)\, \quad \text{with}
    \quad
    p_t^i = \operatorname{softmax}(z_t^i)\in\mathbb{R}^C,
\end{equation*}
where $p_t^i$ denotes the model prediction for input $x_t^i$ at time $t$, and $H(\cdot)$ the entropy of the prediction vector.
Consistent with the motivation above, most TTA methods restrict adaptation to low-entropy samples as these are likely to correspond to mild distributional drift, whereas high-entropy predictions can induce noisy gradients and unstable updates~\cite{niu2023towards,lee2024entropy}.

\subsection{Active Test-Time Adaptation}
\label{sec:prelim-atta}

While unsupervised test-time adaptation can be effective initially, it can become unstable over long test streams due to error accumulation and representation drift~\cite{wang2022continual,niu2022efficient}.
To mitigate this, \emph{active test-time adaptation} (ATTA) adds \emph{limited} supervision~\cite{gui2024active,wang2025effortless}.
In this setting, the model is allowed to query ground-truth labels for a small subset of test samples and incorporate a supervised objective during adaptation.
Let $Q_t \subseteq \mathcal{B}_t$ denote the labeled subset of batch $\mathcal{B}_t$ and let $p_t(x)_y$ denote the predicted probability of the ground-truth class $y$.
Following standard practice, we use the cross-entropy loss on the queried samples,
\begin{equation*}
    L_{\mathrm{sup}}^t
    = \frac{1}{|Q_t|}
      \sum_{(x,y)\in Q_t}
      \big[-\log p_t(x)_y\big],
\end{equation*}
with $L_{\mathrm{sup}}^t = 0$ when $Q_t = \emptyset$.
The supervised signal is combined with entropy minimization via
\begin{equation}
\label{eq:total_loss}
L_{\mathrm{total}}^t
= \lambda_{\mathrm{sup}}\, L_{\mathrm{sup}}^t
+ \lambda_{\mathrm{ent}}\, L_{\mathrm{ent}}^t\,,
\end{equation}
where $\lambda_{\mathrm{sup}}$ and $\lambda_{\mathrm{ent}}$ control the relative contributions of the individual terms. An analysis of which parameters to update during adaptation is
provided in supplementary Appendix~\ref{app:layers}.

\boldpar{Batch-centric label efficiency}
Sparse supervision can stabilize adaptation, but obtaining labels can be expensive~\cite{li2024exploring,chen2024towards}. A central goal of ATTA is therefore to maximize the benefit obtained from each queried label.
Existing ATTA methods typically approach this challenge from a \emph{within-batch} perspective, focusing on reducing the number of annotations required per batch by selecting samples that are either representative or highly learnable.
For example, SimATTA selects samples based on entropy and clustering criteria~\cite{gui2024active}, and EATTA prioritizes samples according to prediction sensitivity to feature perturbations~\cite{wang2025effortless}.
However, this batch-centric view abstracts away the temporal structure of continual test streams. In practice, the contribution of different batches to adaptation can vary substantially over time. By applying supervision uniformly across batches, existing approaches therefore leave open a fundamental question: how should limited supervision be allocated \emph{over time}?

\section{Budgeted Active Test-Time Adaptation}
\label{method}

We address this question by introducing a \emph{budgeted} ATTA setting, where only a limited number of labels can be queried over a long stream. Instead of assuming that every batch is labeled, supervision must be used selectively over time.

\boldpar{Problem setting}
Formally, we consider a stream of test batches $\{\mathcal{B}_t\}_{t=1}^{T}$ and a fixed annotation ratio $r \in [0,1]$ that specifies the fraction of batches for which supervision may be requested. At each time step $t$, the adaptation procedure decides whether to request supervision for the current batch $\mathcal{B}_t$.
This decision is represented by a binary action $a_t \in \{0,1\}$, where $a_t = 1$ indicates that batch $\mathcal{B}_t$ is selected for supervision and $a_t = 0$ otherwise. If $a_t = 1$, exactly one label is queried from $\mathcal{B}_t$, yielding a labeled set $Q_t \subseteq \mathcal{B}_t$ with $|Q_t| = 1$; if $a_t = 0$, then $Q_t = \emptyset$ 
.
The total annotation budget is enforced by the constraint
\begin{equation*}
    \sum_{t=1}^{T} |Q_t| \le \lfloor rT \rfloor.
\end{equation*}
This setting couples two aspects of supervision usage: allocating limited supervision over time and using each queried label as effectively as possible.

To address both aspects, we introduce \emph{WISE-ATTA}, which consists of
(i) a budget-paced batch selection strategy that decides \emph{when} to request supervision, and
(ii) a drift-based single-sample selection rule that determines \emph{what} to label within a selected batch.
The two decisions serve distinct roles: batch selection identifies a reliable adaptation regime, while sample selection identifies where the model remains responsive to correction.
Below, we describe each component in turn and then specify the adaptation update.

\subsection{Batch Selection}
\label{sec:batch-level-selection}

We start by considering how supervision should be allocated across the stream.
The goal is to identify batches where supervision is likely to yield meaningful updates, using only signals available online, without knowledge of future shifts.

\boldpar{Batch utility}
To decide whether a batch warrants supervision, we need an unsupervised
signal, since labels are unavailable at this stage.
We use prediction entropy as a lightweight indicator of the model's
current adaptation regime. Low entropy corresponds to a concentrated
prediction, indicating that the model assigns relatively high probability
to a small number of classes.
Thus, the fraction of low-entropy predictions measures how much of the
current batch lies in a comparatively confident prediction region.
Under the standard assumption that confidence correlates with
correctness, we use this quantity as a proxy for batch-level adaptation
reliability.
Importantly, confidence is used here to decide \emph{when} to supervise,
rather than \emph{which} sample to label. Unlike conventional uncertainty
sampling, the batch-level objective is to identify a regime in which a
sparse corrective label can be incorporated reliably into the ongoing
adaptation process.
Formally, for each incoming batch
$\mathcal{B}_t=\{x_t^i\}_{i=1}^{n}$, we define
\begin{equation}
\label{eq:utility}
    s_t
    = \sum_{i=1}^{n}
    \mathbb{I}\!\left[
        H(p_t^i) < \tau_{\mathrm{ent}}
    \right],
\end{equation}
where
$\tau_{\mathrm{ent}}$ is a fixed threshold.
Intuitively, $s_t$ measures the confident mass of the current batch and
serves as a proxy for batch-level adaptation reliability.
Alternative definitions of the batch-level utility proxy are analyzed and evaluated in supplementary Appendix~\ref{sec:utility}.

\boldpar{Utility-based allocation over time}
Given the per-batch utility score $s_t$, we still need a policy that decides \emph{when} to spend supervision over the course of the stream. Two constraints shape this decision.
First, supervision is limited, so labels should be allocated to batches that are informative \emph{relative to recent observations}.
Recent batches provide a natural reference for the model's current adaptation regime: both the model parameters and the input distribution evolve over time, so a utility score that appears high in absolute terms may be typical relative to the current stream state.
Second, in realistic online deployments the total stream length is often unknown, ruling out policies that explicitly plan over the remaining horizon.

To address the first constraint, we maintain a sliding history buffer $\mathcal{H}$ of the most recent $W$ batch utility scores and request supervision whenever the current score exceeds an \textit{adaptive quantile threshold} $\tau_t$:
\begin{equation*}
\label{eq:threshold}
    a_t = \mathbb{I}[s_t \ge \tau_t]\,, 
    \qquad
    \tau_t = \mathrm{Quantile}_{\,1-\tilde{r}_t}(\mathcal{H})\,.
\end{equation*}
The quantile level is governed by an \emph{effective annotation rate} $\tilde{r}_t \in [0,1]$, which controls how permissive the policy currently is; a larger $\tilde{r}_t$ lowers the threshold and admits more batches, while a smaller $\tilde{r}_t$ tightens it.

To address the second constraint, we set $\tilde{r}_t$ via a pacing controller that does not require knowing the stream length. Let $r \in [0,1]$ denote the desired long-run annotation ratio and $u_t = \sum_{k=1}^{t-1} a_k$ the number of labels used before time $t$. We define the \emph{budget debt} as
\begin{equation*}
\label{eq:debt}
    d_t = rt - u_t\,,
\end{equation*}
i.e., the gap between the cumulative usage targeted under rate $r$ and the labels actually spent so far. The effective rate is then
\begin{equation*}
    \tilde{r}_t = r + \frac{d_t}{H_c}\,,
\end{equation*}
where $H_c > 0$ is a local correction horizon. We clip $\tilde{r}_t$ to $[0,1]$ to keep it valid as a quantile level.
Intuitively, when the policy has used labels too slowly ($d_t > 0$), $\tilde{r}_t$ increases and the threshold becomes more permissive; when labels have been spent too quickly ($d_t < 0$), $\tilde{r}_t$ decreases and the threshold tightens.

\boldpar{Practical refinements}
We add two refinements to this strategy.
First, at the start of the stream when the history buffer is still being populated ($|\mathcal{H}| < M$), there are too few past scores to estimate a reliable quantile threshold.
During this warmup phase, we instead sample $a_t \sim \mathrm{Bernoulli}(r)$, matching the target annotation rate in expectation while $\mathcal{H}$ fills up.
Second, to prevent short-term fluctuations from causing persistent under-utilization of the budget, we apply a rate-floor correction.
Let $u_t$ denote the number of labels used up to time $t$ and let $u_t^\star = rt$ be the target usage. We force annotation whenever $u_t + \delta < u_t^\star$, where $\delta \ge 0$ is a slack parameter that tolerates small transient deviations.
We ablate these design choices in supplementary Appendix~\ref{app:hyper_sensitivity}.

\subsection{Sample Selection}
\label{sec:batch-sample-selection}

Once a batch is selected for supervision, we must decide which sample to annotate. A natural
choice would be prediction entropy, but it is poorly suited here as it is a static criterion;
a snapshot of confidence at a single point in time and tells us nothing about whether that
confidence is the result of ongoing adaptation or a stable end state.
For this, we require a
temporal criterion that captures how a sample’s predictions evolve under adaptation, and
thereby identifies the samples for which a label is informative. 
We discuss the complementary roles of entropy (batch level) and drift (sample level) in more detail in supplementary Appendix~\ref{app:batch_vs_sample}.

\boldpar{Prediction drift as an adaptation signal}
We capture this by examining how model predictions evolve and focus on samples whose predictions change consistently under ongoing adaptation. Such prediction drift indicates that the model is actively adjusting for these inputs, but has not yet stabilized. Supervising samples in this regime can be particularly effective as the model is receptive to correction, yet sufficiently aligned for the label to propagate reliably.

\boldpar{Anchor model}
To this end, we maintain an exponential moving average (EMA) of the model as a temporally smoothed reference. Let $f_{\theta_f}$ and $h_{\theta_c}$ denote the current feature extractor and classifier, and let
$\bar{f}_{\bar{\theta}_f}$ and $\bar{h}_{\bar{\theta}_c}$ denote their EMA counterparts (the \emph{anchor} model).
For an incoming batch $\mathcal{B}_t=\{x_t^i\}_{i=1}^n$, we compute class-probability predictions from the current model and the anchor, denoted $p_t^i$ and $\bar{p}_t^i$, respectively, and define the prediction drift as
\begin{equation*}
d_t^i = \|p_t^i - \bar{p}_t^i\|_2 \, .
\end{equation*}
When $\mathcal{B}_t$ is selected for supervision, we query the label of the sample with the largest drift:
\begin{equation*}
i_t^\star = \arg\max_{1\le i\le n} d_t^i,\qquad Q_t=\{(x_t^{i_t^\star}, y_t^{i_t^\star})\}.
\end{equation*}
Finally, we update the anchor parameters after each batch with the current model with momentum $\mu \in (0,1)$:
\begin{align}
\label{pds-eq}
    \bar{\theta}_f &\leftarrow \mu \bar{\theta}_f + (1 - \mu)\,\theta_f, \\
    \bar{\theta}_c &\leftarrow \mu \bar{\theta}_c + (1 - \mu)\,\theta_c.
\end{align}
The drift admits a local trajectory-sensitivity interpretation.
Let $\Delta_t=\theta_t-\bar{\theta}_t$ and
$J_t^i=\nabla_\theta p_\theta(x_t^i)|_{\theta=\bar{\theta}_t}$.
A first-order expansion gives
\begin{equation}
    p_{\theta_t}(x_t^i)-p_{\bar{\theta}_t}(x_t^i)
    \approx J_t^i\Delta_t,
\end{equation}
and hence
\begin{equation}
    (d_t^i)^2
    \approx
    \Delta_t^\top (J_t^i)^\top J_t^i\Delta_t.
    \label{eq:drift-interpretation}
\end{equation}
Thus, drift measures how responsive a sample's prediction is along the
model's recent adaptation direction, rather than its uncertainty at a
single instant.
\cref{alg:pubs} in supplementary Appendix~\ref{app:implementation} summarizes the full procedure.

\begin{figure*}[t]
  \centering
  \includegraphics[width=\textwidth]{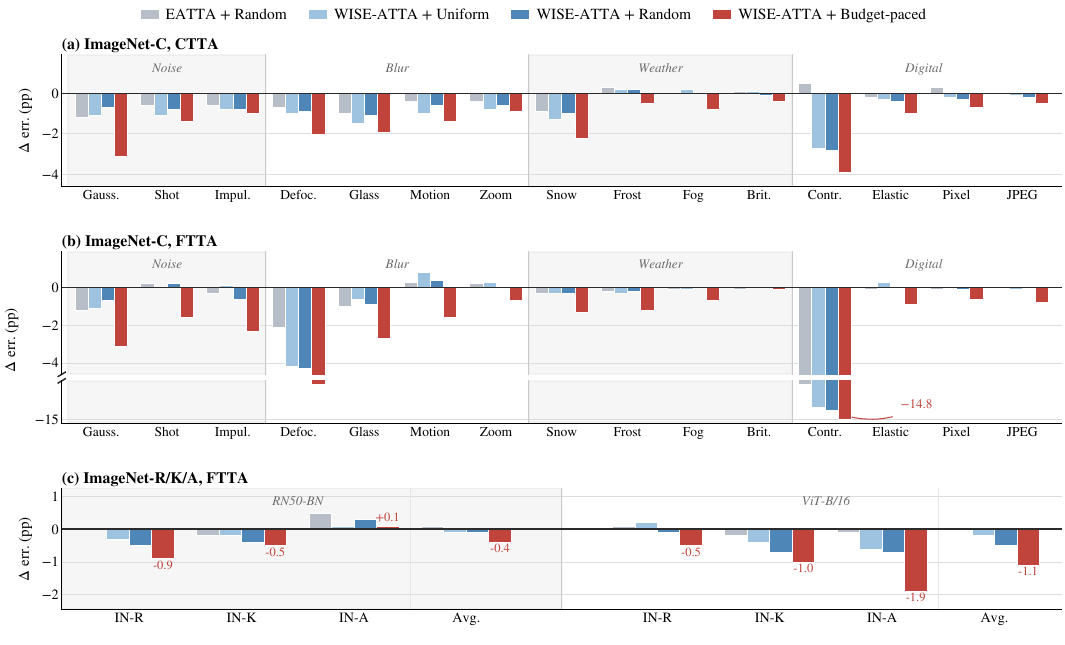}
\caption{\textbf{Batch selection at 0.5 labels/batch.}
Error change relative to EATTA~\cite{wang2025effortless} with uniform batch allocation; lower is better.
\textbf{(a)} ImageNet-C under CTTA, \textbf{(b)} ImageNet-C under FTTA (both ResNet-50), and \textbf{(c)} ImageNet-R/K/A under FTTA.
Our budget-paced selection consistently outperforms uniform and random allocation.
Full results are in supplementary Appendix~\ref{app:batch_selection_results}.}
  \label{fig:batchsel}
\end{figure*}

\section{Evaluation}
\label{sec:evaluation}

We evaluate WISE-ATTA in the budgeted ATTA setting from several complementary perspectives.
We first isolate the contribution of budget-paced batch selection under a fixed annotation budget (\cref{sec:batch-selection-analysis}).
We then evaluate the effectiveness of our drift-based sample selection by comparing against prior active TTA methods under both matched and larger annotation budgets (\cref{sec:sample-selection-analysis}).
Next, we study performance across a range of label budgets (\cref{sec:label_ratio}) and analyze how WISE-ATTA allocates supervision over time and across samples (\cref{sec:label_utilization,sec:selected_sample}).
We further examine robustness to delayed label availability in supplementary Appendix~\ref{app:annotation_delay} and to different online batch sizes in supplementary Appendix~\ref{app:batchsize}.
Unless stated otherwise, we use a batch size of 64.
Additional implementation details, algorithmic settings, datasets, models, and reproducibility information are provided in supplementary Appendices~\ref{app:implementation} and~\ref{app:experimental_details}.

\subsection{Analysis of Batch Selection Strategies}
\label{sec:batch-selection-analysis}
We begin by asking whether our budget-paced batch selection criterion in fact identifies the important batches to annotate; those whose supervision yields the largest adaptation gains per labeled sample. 
Answering this requires isolating batch selection from the orthogonal axis of which samples within a batch get annotated, since the two are easily confounded. Strong sample selector can mask a weak batch selector, and vice versa.
To this end, we hold the sample-selection method fixed and vary only the batch selection strategy. We further compare the budget-paced selection against two simple strategies: \textsc{Uniform}, which spreads the annotation budget evenly across batches, and \textsc{Random}, which allocates it stochastically. 
We repeat this comparison under two sample selection methods: EATTA~\cite{wang2025effortless} and our drift-based selector.

We run these experiments on ImageNet-C, R, K, and A. 
For ImageNet-C, we evaluate under two standard test-time adaptation protocols. First, \emph{fully test-time adaptation} (FTTA)~\cite{wang2020tent}, where the model is reset to the pretrained checkpoint whenever the corruption type changes. Second, \emph{continual test-time adaptation} (CTTA)~\cite{wang2022continual}, where adaptation proceeds over a single continuous stream without resets.

\begin{table*}[t]
\centering
\caption{\textbf{ImageNet-C} error (\%) under CTTA (top) and FTTA (bottom).
\#Labels denotes the average number of queried labels per batch; $\mathcal{BFS}$ denotes the replay buffer size when used.
$^\dagger$ indicates methods that require a teacher model.
Our methods are highlighted in gray.}
\label{tab:imagenet-c}
\text{\tiny \textbf{CTTA Setting}}
\begin{threeparttable}
\resizebox{\linewidth}{!}{
\begin{tabular}{c l | ccc | cccc | cccc | cccc | c}
\toprule
\textbf{\# Labels} & \textbf{Method}
& \multicolumn{3}{c}{Noise}
& \multicolumn{4}{c}{Blur}
& \multicolumn{4}{c}{Weather}
& \multicolumn{4}{c}{Digital}
& \textbf{Avg. Err.} \\
\cmidrule(lr){3-5}\cmidrule(lr){6-9}\cmidrule(lr){10-13}\cmidrule(lr){14-17}
&
& Gauss. & Shot & Impul.
& Defoc. & Glass & Motion & Zoom
& Snow & Frost & Fog & Brit.
& Contr. & Elastic & Pixel & JPEG
& \\
\midrule
\multirow{4}{*}{\rotatebox{90}{Non-Active}}
& TENT~\cite{wang2020tent} & 70.8 & 63.9 & 64.9 & 75.7 & 75.1 & 72.3 & 65.3 & 72.8 & 75.7 & 68.4 & 56.0 & 84.0 & 73.1 & 70.7 & 74.9 & 70.9 \\
& CoTTA~\cite{wang2022continual} & 78.2 & 68.6 & 64.3 & 75.2 & 71.5 & 69.6 & 67.5 & 72.0 & 71.4 & 67.2 & 62.3 & 73.5 & 69.4 & 66.8 & 68.6 & 69.8 \\
& SAR~\cite{niu2023towards} & 70.0 & 62.2 & 62.9 & 73.0 & 70.1 & 65.7 & 58.0 & 63.8 & 64.1 & 53.6 & 42.0 & 68.3 & 53.8 & 50.3 & 53.2 & 60.7 \\
& ETA~\cite{niu2022efficient} & 65.2 & 59.5 & 61.1 & 70.0 & 69.0 & 63.8 & 57.3 & 58.9 & 60.8 & 48.7 & 39.5 & 58.2 & 48.6 & 45.2 & 48.1 & 56.9 \\
\midrule

\multirow{2}{*}{3} 
& SimATTA~\cite{gui2024active} ($\mathcal{BFS}=300$) & 65.3 & 59.2 & 60.4 & 68.0 & 65.0 & 58.4 & 53.0 & 54.9 & 57.5 & 45.9 & 38.0 & 56.8 & 48.3 & 43.8 & 47.2 & 54.8 \\
& HILTTA~\cite{li2024exploring} & 65.1 & 57.6 & 58.5 & 65.5 & 63.1 & 56.9 & 51.7 & 54.4 & 56.1 & 46.1 & 36.9 & 55.4 & 47.2 & 42.7 & 46.2 & 53.7 \\
\midrule
\multirow{2}{*}{1}
& EATTA~\cite{wang2025effortless}  & 64.9 & 57.5 & 58.0 & 67.1 & 65.2 & 57.7 & 53.7 & 55.3 & 58.3 & 46.5 & 37.7 & 59.2 & 47.9 & 43.6 & 46.7 & 54.6 \\
& \gcell{WISE-ATTA} & \gcell{64.1} & \gcell{57.2} & \gcell{57.6} & \gcell{67.3} & \gcell{65.1} & \gcell{57.0} & \gcell{52.5} & \gcell{53.1} & \gcell{56.6} & \gcell{43.9} & \gcell{36.4} & \gcell{54.9} & \gcell{46.3} & \gcell{41.8} & \gcell{45.1} & \gcell{\textbf{53.3}} \\
\midrule
\end{tabular}
}
\end{threeparttable}

\text{\tiny \textbf{FTTA Setting}}
\begin{threeparttable}
\resizebox{\linewidth}{!}{
\begin{tabular}{c l | ccc | cccc | cccc | cccc | c}
\toprule
\textbf{\# Labels} & \textbf{Method}
& \multicolumn{3}{c}{Noise}
& \multicolumn{4}{c}{Blur}
& \multicolumn{4}{c}{Weather}
& \multicolumn{4}{c}{Digital}
& \textbf{Avg. Err.} \\
\cmidrule(lr){3-5}\cmidrule(lr){6-9}\cmidrule(lr){10-13}\cmidrule(lr){14-17}
&
& Gauss. & Shot & Impul.
& Defoc. & Glass & Motion & Zoom
& Snow & Frost & Fog & Brit.
& Contr. & Elastic & Pixel & JPEG
& \\
\midrule

\multirow{4}{*}{\rotatebox{90}{Non-Active}}
& TENT~\cite{wang2020tent}
& 70.8 & 69.0 & 70.0 & 71.9 & 72.0 & 58.3 & 50.7
& 52.6 & 58.5 & 42.4 & 32.7
& 70.3 & 45.3 & 41.5 & 47.6
& 56.9 \\
& CoTTA~\cite{wang2022continual}
& 78.2 & 77.8 & 77.2 & 81.8 & 77.8 & 63.8 & 53.2
& 57.6 & 60.4 & 44.1 & 32.8
& 73.5 & 48.9 & 43.0 & 52.6
& 61.5 \\
& SAR~\cite{niu2023towards}
& 69.9 & 69.2 & 69.1 & 71.2 & 71.7 & 57.9 & 50.8
& 52.9 & 57.8 & 42.5 & 32.7
& 62.3 & 45.7 & 41.7 & 47.8
& 56.2 \\
& ETA~\cite{niu2022efficient}
& 65.2 & 62.4 & 63.5 & 66.8 & 66.6 & 52.6 & 47.2
& 48.4 & 53.8 & 40.2 & 32.2
& 54.8 & 42.3 & 39.2 & 45.0
& 52.0 \\
\midrule

\multirow{1}{*}{-}
& CEMA$^\dagger$~\cite{chen2024towards} ($\mathcal{BFS}=300$)
& 64.9 & 69.1 & 62.7 & 66.7 & 66.5 & 52.7 & 48.4
& 48.6 & 54.6 & 40.6 & 33.5
& 57.4 & 43.2 & 40.1 & 45.1
& 52.9 \\
\multirow{2}{*}{3}
& SimATTA~\cite{gui2024active} ($\mathcal{BFS}=300$)
& 67.4 & 63.7 & 65.5 & 68.1 & 66.8 & 55.3 & 49.7
& 51.3 & 55.9 & 42.7 & 33.7
& 56.3 & 45.4 & 41.7 & 47.4
& 54.1 \\
& HILTTA~\cite{li2024exploring}
& 65.0 & 63.1 & 64.5 & 66.1 & 66.3 & 54.4 & 48.4
& 49.8 & 54.8 & 41.4 & 32.5
& 55.6 & 43.6 & 40.5 & 45.8
& 52.8 \\
\midrule

\multirow{2}{*}{1}
& EATTA~\cite{wang2025effortless}
& 64.9 & 62.4 & 63.9 & 68.0 & 66.9 & 51.6 & 47.7
& 47.9 & 54.2 & 40.2 & 31.7
& 64.0 & 42.7 & 39.1 & 44.9
& 52.7 \\

& \gcell{WISE-ATTA}
& \gcell{64.1} & \gcell{61.5} & \gcell{62.5} & \gcell{66.9} & \gcell{65.9} & \gcell{50.8} & \gcell{47.4}
& \gcell{47.5} & \gcell{53.8} & \gcell{39.7} & \gcell{32.0}
& \gcell{56.8} & \gcell{42.3} & \gcell{38.9} & \gcell{44.4}
& \gcell{\textbf{51.6}} \\
\midrule

\end{tabular}
}
\end{threeparttable}
\end{table*}
\begin{table*}[!t]
\caption{Generalization under natural distribution shifts: FTTA error (\%) on ImageNet-R/K/A with RN50-BN and ViT-B-16.}
\label{tab:imagenet-variants}
\centering
\resizebox{0.9\linewidth}{!}{
\begin{tabular}{@{} c l | cc | cc | cc || cc}
\toprule
\textbf{\# Labels} & \textbf{Method}
& \multicolumn{2}{c}{ImageNet-R} 
& \multicolumn{2}{c}{ImageNet-K}
& \multicolumn{2}{c}{ImageNet-A}
& \multicolumn{2}{c}{\textbf{Avg. Error}} \\
\cmidrule(lr){3-4} \cmidrule(lr){5-6} \cmidrule(lr){7-8} \cmidrule(lr){9-10}
& 
& RN50-BN & ViT-B-16
& RN50-BN & ViT-B-16
& RN50-BN & ViT-B-16
& RN50-BN & ViT-B-16 \\
\midrule

\multirow{4}{*}{\rotatebox{90}{Non-Active}}
& TENT~\cite{wang2020tent}
& 57.8 & 53.4 & 69.5 & 65.6 & 99.9 & 77.4 & 75.7 & 65.5 \\
& CoTTA~\cite{wang2022continual}
& 57.3 & 55.4 & 69.9 & 98.2 & 99.8 & 79.3 & 75.7 & 77.6 \\
& SAR~\cite{niu2023towards}
& 57.2 & 48.8 & 68.5 & 70.4 & 99.9 & 74.9 & 75.2 & 64.7 \\
& ETA~\cite{niu2022efficient}
& 54.0 & 48.8 & 64.3 & 59.4 & 99.8 & 75.9 & 72.7 & 61.4 \\
\midrule

\multirow{1}{*}{-}
& CEMA$^\dagger$~\cite{chen2024towards}
& 51.4 & 44.6 & 65.6 & 60.0 & 97.7 & 72.9 & 71.6 & 59.2 \\
\multirow{2}{*}{3}
& SimATTA~\cite{gui2024active}
& 51.3 & 45.1 & 64.0 & 57.2 & 97.2 & 72.4 & 70.8 & 58.2 \\
& HILTTA~\cite{li2024exploring}
& 52.6 & 43.9 & 63.3 & 58.1 & 98.3 & 72.2 & 71.4 & 58.1 \\
\midrule

\multirow{2}{*}{1}
& EATTA~\cite{wang2025effortless}
& 52.8 & 44.3 & 64.1 & 58.2 & 99.1 & 71.5 & 72.0 & 58.0 \\

& \cellcolor{gray!12}WISE-ATTA
& \cellcolor{gray!12}51.2 & \cellcolor{gray!12}42.6
& \cellcolor{gray!12}63.8 & \cellcolor{gray!12}57.2
& \cellcolor{gray!12}97.7 & \cellcolor{gray!12}69.9
& \cellcolor{gray!12}70.9 & \cellcolor{gray!12}\textbf{56.6} \\
\midrule

\end{tabular}
}

\end{table*}

\boldpar{Results}
\Cref{fig:batchsel} reports the error change relative to the
\textsc{EATTA + Uniform} baseline under a fixed budget of $0.5$ labels
per batch. Two findings stand out.
First, our budget-paced selection generally yields the largest error
reductions across ImageNet-C under both CTTA and FTTA, with particularly
large gains on several corruptions.
Second, under matched uniform or random batch allocation, our drift-based
sample selector already improves over EATTA, while combining it with
budget-paced selection provides further gains.
The same trend holds on ImageNet-R/K/A across both backbones, with average
improvements of $0.4$ and $1.1$ points for RN50-BN and ViT-B-16,
respectively.
Overall, the results show that batch- and sample-level selection provide
complementary benefits under the same annotation budget.

\subsection{Effectiveness of Drift-Based Sample Selection}
\label{sec:sample-selection-analysis}

Having established that budget-paced batch selection helps, we next ask the complementary question: given a batch, does WISE-ATTA pick the right samples to annotate?
This matters because batch and sample selection are independent levers; gains from the former say nothing about the quality of the latter.
To answer this, we compare WISE-ATTA against two families of baselines. The first is active TTA methods, which, like WISE-ATTA, choose which samples to annotate within the test stream: SimATTA~\cite{gui2024active} and HILTTA~\cite{li2024exploring} (both querying three labels per batch), CEMA~\cite{chen2024towards} (which additionally requires a replay buffer and a strong teacher), and the recent single-label method EATTA~\cite{wang2025effortless}. 
The second family are non-active, fully unsupervised TTA methods; that is, TENT~\cite{wang2020tent}, CoTTA~\cite{wang2022continual}, SAR~\cite{niu2022efficient}, and ETA~\cite{lee2024entropy}. Including these is important for context as they establish the no-supervision floor. The gap between them and the active methods quantifies how much value labels can provide, and the gap toWISE-ATTA quantifies how much of that can be captured by WISE-ATTA.

\boldpar{Results}
\Cref{tab:imagenet-c} reports ImageNet-C results under both CTTA and FTTA protocols, and \cref{tab:imagenet-variants} reports results on ImageNet-R/K/A.
On ImageNet-C, WISE-ATTA achieves the lowest average error in both settings (CTTA: 53.3\%; FTTA: 51.6\%), improving over EATTA by $1.3$ and $1.1$ points respectively despite using the same $1$-label-per-batch budget. Notably, WISE-ATTA also outperforms the higher-budget HILTTA ($3$ labels per batch) by $0.4$ points under CTTA and $1.2$ points under FTTA, indicating that better sample selection can substantially offset more annotation effort.
The same pattern holds under natural distribution shifts in \cref{tab:imagenet-variants}. WISE-ATTA achieves the best average error for both models (RN50-BN: 70.9\%; ViT-B-16: 56.6\%), again improving over EATTA at the same budget and matching or exceeding the $3$-label active baselines.

Together, this indicates that prediction drift is a stronger per-sample selection criterion than the entropy- and confidence-based signals used in prior active TTA, and that the gain transfers across both adaptation protocols and shift types.

\subsection{Performance evaluation under varying label ratio}
\label{sec:label_ratio}

Thus far, we have focused on two annotation ratios ($r=0.5$ and $r=1$). 
We now broaden the view and ask how WISE-ATTA performs, and how it compares to baselines, across a wide range of $r$.
Sweeping $r$ lets us characterize how WISE-ATTA's advantage scales with supervision.
To isolate the effect of the batch selection strategy, we again compare against \textsc{Uniform} and \textsc{Random} batch selection. For a fair comparison, we enforce full budget utilization for all approaches by labeling all remaining batches once the remaining budget suffices for the remaining batches.

\begin{figure}[h]
  \centering
  \includegraphics[width=0.48\linewidth]{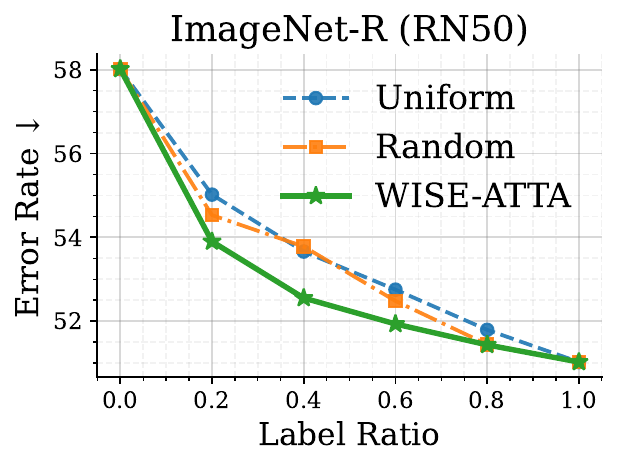}
  \includegraphics[width=0.48\linewidth]{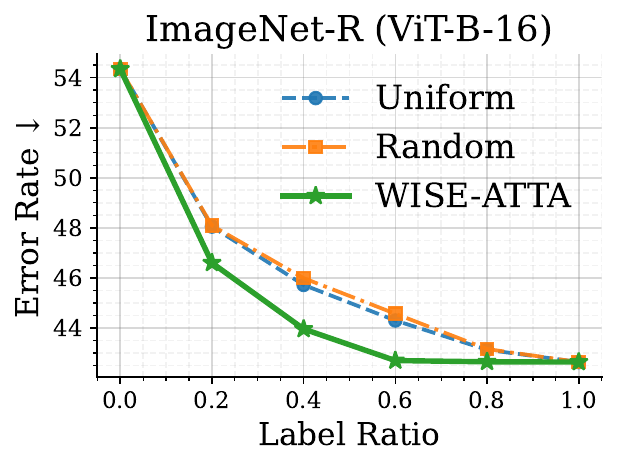}
  \includegraphics[width=0.48\linewidth]{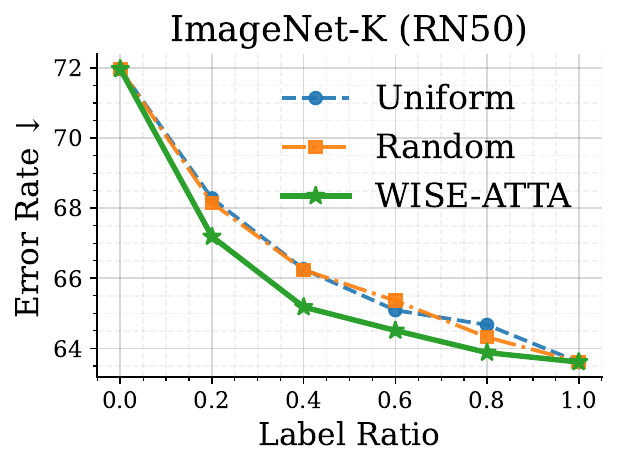}
  \includegraphics[width=0.48\linewidth]{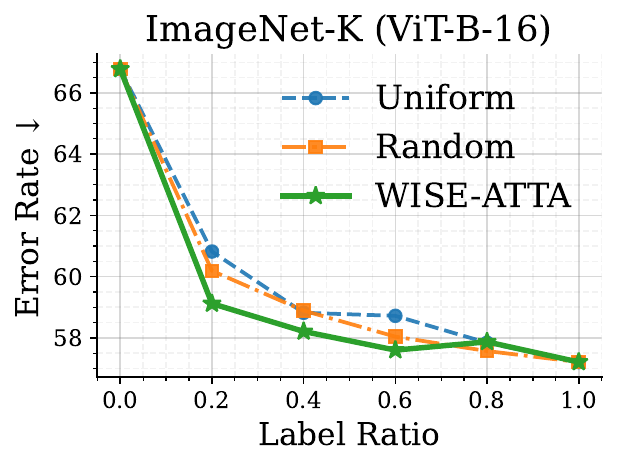}
  \vspace{-0.7em}
\caption{Performance across varying label ratios on ImageNet-R/K}
  \label{fig:label_ratio}
\end{figure}

\boldpar{Results}
\cref{fig:label_ratio} shows results on ImageNet-R and ImageNet-K.
Across all settings, WISE-ATTA outperforms both \textsc{Uniform} and \textsc{Random} selection, which behave similarly across label ratios and fail to exploit differences in batch utility.
The gains are most pronounced in the low-budget regime: increasing $r$ from $0$ to $0.2$ improves performance by 4.12 points on ImageNet-R with RN50-BN ($58.02 \rightarrow 53.90$) and by 7.66 points on ImageNet-K with ViT-B-16 ($66.78 \rightarrow 59.12$).
As the budget increases, WISE-ATTA approaches the performance of labeling every batch; at $r=0.6$, it nearly matches performance at $r=1.0$ while using substantially fewer labels.

\subsection{Label Utilization over time}
\label{sec:label_utilization}

\begin{figure}[h]
  \centering
  \includegraphics[width=0.48\linewidth]{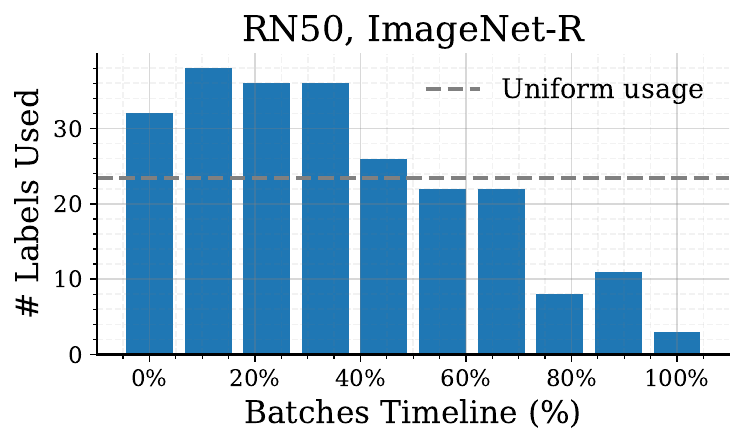}
  \includegraphics[width=0.48\linewidth]{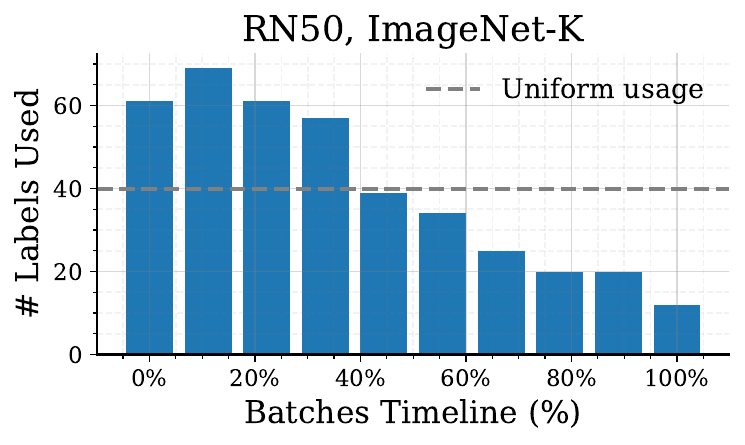}

  \includegraphics[width=0.48\linewidth]{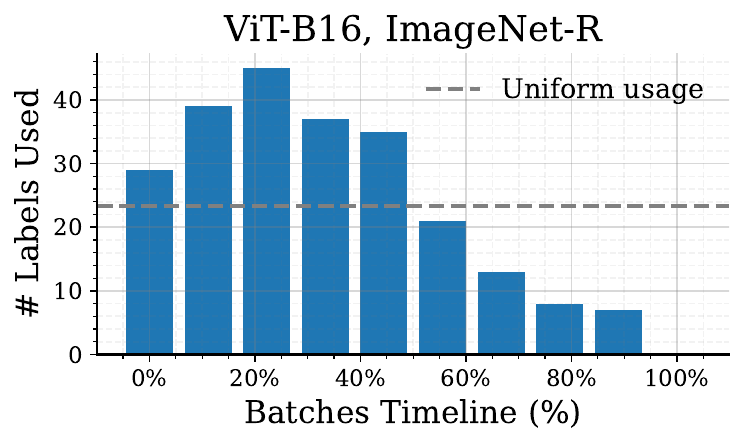}
  \includegraphics[width=0.48\linewidth]{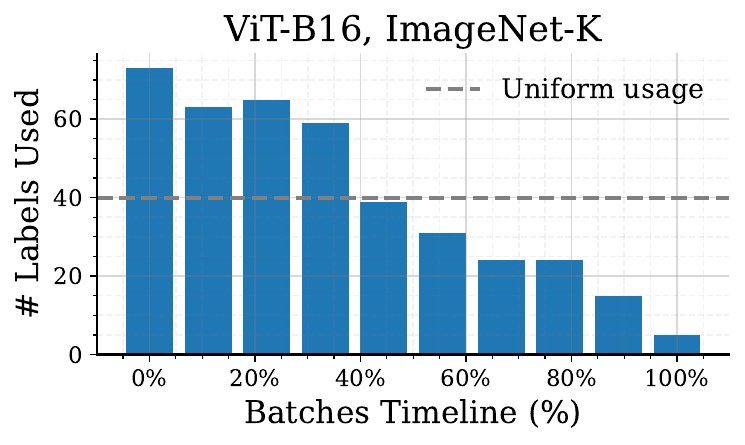}
    \vspace{-0.6em}

  \caption{Label utilization over the test-time stream.}
  \label{fig:label_usage_bins}
\end{figure}

To understand why WISE-ATTA is effective, we next examine how it allocates its label budget over the test stream. Visualizing the allocation tells us where in the stream WISE-ATTA chooses to spend its budget
We partition the stream into ten equal time bins and report the number of labeled batches per bin (\cref{fig:label_usage_bins}). The dashed line indicates uniform allocation as a reference.

\boldpar{Results} Across datasets and models, WISE-ATTA allocates labels non-uniformly, with higher density early in the stream.
This behavior is desirable: when a new distribution shift is first encountered, the model is least aligned with the target domain and supervision provides the largest marginal benefit.
The front-loading effect is especially pronounced on ImageNet-K, where label usage steadily decreases over time. On ImageNet-R, label allocation is broader, peaking in the early-to-mid portion of the stream ($\approx10-40\,\%$).
Importantly, WISE-ATTA does not exhaust the budget immediately and continues to allocate labels near the end of evaluation, allowing it to react to high-utility batches throughout.

\subsection{Selected Sample Analysis}
\label{sec:selected_sample}

\begin{figure}[t]
    \centering

    \includegraphics[width=0.48\linewidth]{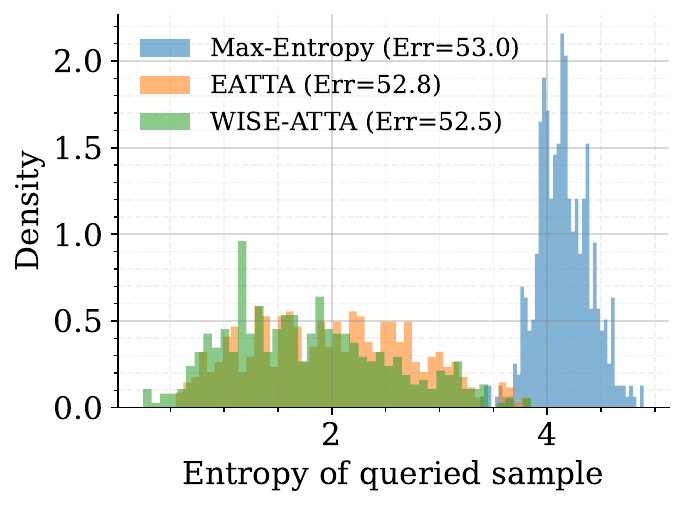}
    \hfill
    \includegraphics[width=0.48\linewidth]{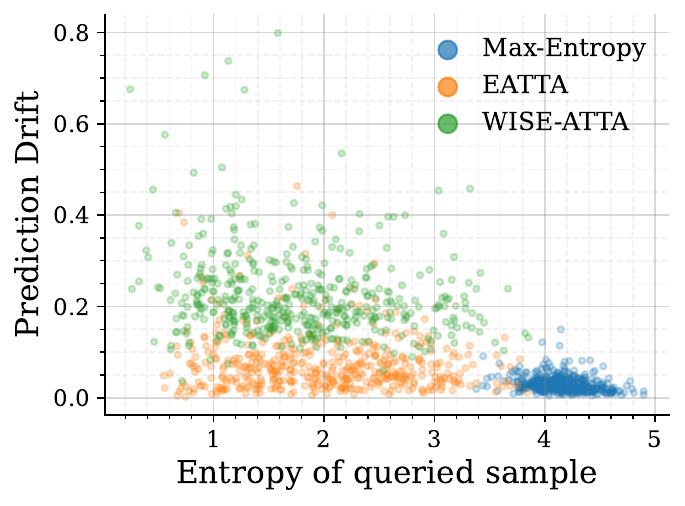}

    \vspace{-0.4em}

    \makebox[0.48\linewidth][c]{\footnotesize (a)}
    \hfill
    \makebox[0.48\linewidth][c]{\footnotesize (b)}

    \vspace{-0.5em}

\caption{\textbf{Query behavior on ImageNet-R.}
(a) Entropy distribution of samples queried by Max-Entropy~\cite{wang2014new}, EATTA~\cite{wang2025effortless}, and WISE-ATTA.
(b) Prediction drift (vs.\ EMA anchor) versus entropy.
Additional results and discussion are in supplementary Appendix~\ref{app:query_behavior}.}
    
    \label{fig:motivation_plots_1}
\end{figure}

We now zoom in from when WISE-ATTA spends its budget to which samples it spends it on.
Our sample selection is based on the observation to select samples whose predictions change in a consistent and directional manner.
\cref{fig:motivation_plots_1} contrasts samples queried by Max-Entropy, EATTA, and WISE-ATTA.

Max-Entropy focuses on the high-entropy tail, which can contain overly ambiguous samples that are difficult to exploit with a single labeled update, leading to the weakest performance.
EATTA shifts toward lower-to-moderate entropy but selects samples with relatively small prediction drift, indicating that supervision is often spent in already stable regions.
In contrast, WISE-ATTA avoids the highest-entropy extremes while prioritizing samples with higher drift relative to its EMA anchor, capturing samples that are still adapting and thus responsive to corrective supervision.

\section{Discussion}
\label{sec:discussion}
Our findings highlight a practical lesson for test-time adaptation: if supervision is scarce, \emph{when} labels are used can be as important as \emph{which} samples are labeled.
Below, we take a broader perspective on this and discuss assumptions, limitations, and future work.

\boldpar{Adaptation under limited resources}
WISE-ATTA explicitly budgets supervision, but still updates the model at every time step. In compute constrained settings, it may also be necessary to budget \emph{updates} themselves.
Skipping updates does not merely reduce computation; it can also change the balance between
unsupervised and supervised correction. In supplementary Appendix~\ref{app:skip_updates}, we adapt WISE-ATTA to skip unsupervised updates on non-selected batches, and find that it remains competitive despite reduced update frequency.
This suggests the need for principled criteria to decide when entropy minimization should be relied upon versus when supervised correction should dominate. %

\boldpar{Data and shift structure}
We consider test streams in which each batch is dominated by a single distribution shift. This is an oversimplification for many scenarios and a natural extension is \emph{heterogeneous} batches where multiple shift sources co-occur within the same batch or within short temporal windows. In such settings, batch utility may need to reflect coverage across multiple modes, rather than short-term adaptation potential alone.

\boldpar{Resilience to stale supervision}
Our delay analysis (\cref{app:annotation_delay}) shows that ATTA methods degrade sharply when labels arrive late. This points to delay-aware extensions like recency-weighted updates, latency-aware batch selection, or query strategies that anticipate the model's future state as a necessary step toward deployment-ready active~TTA.

\section{Related Work}

Models deployed in the wild often face distribution shift, leading to significant performance degradation~\cite{hendrycks2019benchmarking,recht2019imagenet,ovadia2019can}.

\boldpar{Test-time adaptation}
To address this without frequent retraining, prior work has explored \emph{test-time training} (TTT) and \emph{test-time adaptation} (TTA).
TTT optimizes an auxiliary self-supervised objective learned on the source domain at deployment and is therefore not fully off-the-shelf~\cite{sun2020test,gidaris2018unsupervised}.
In contrast, TTA adapts a pretrained model directly on the unlabeled test stream using unsupervised objectives such as entropy minimization or self-training~\cite{wang2020tent,wang2022continual,liang2020we}.
While effective initially, unsupervised TTA can become unstable under long or severe shifts due to error accumulation, catastrophic forgetting, and drift in feature statistics~\cite{chen2022contrastive,niu2022efficient,wang2022continual,hu2021mixnorm,su2024towards}.
To mitigate these issues, prior work filters unreliable samples~\cite{lee2024entropy,niu2022efficient}, refines pseudo-labels~\cite{chen2022contrastive}, adds regularization to reduce forgetting~\cite{brahma2023probabilistic,wang2022continual}, impose prototype-based constraints~\cite{dobler2023robust,jang2022test,wang2025decoupled}, or corrects feature statistics via running normalization estimates~\cite{hong2023mecta,mirza2022norm,yuan2023robust}.
Nevertheless, purely unsupervised objectives remain vulnerable to confirmation bias and error accumulation, motivating occasional supervision at test time~\cite{gui2024active,wang2025effortless}.

\boldpar{Active test-time adaptation}
Active learning (AL) studies how to query labels under a limited budget ~\cite{li2024survey}, using strategies as uncertainty, disagreement, expected model change, or diversity~\cite{wang2014new,seung1992query,yoo2019learning,sener2017active}.
Active test-time adaptation (ATTA) brings these ideas to the test-time setting, using sparse supervision to stabilize and guide online adaptation~\cite{gui2024active,wang2025effortless}.
Previous work in this area has focused on \emph{what} to label within each batch, via uncertainty-based querying~\cite{gui2024active}, online model selection~\cite{li2024exploring}, or teacher-guided distillation~\cite{chen2024towards}, with recent methods reducing supervision to even a single label per batch~\cite{wang2025effortless}.
However, most ATTA methods assume a fixed querying cadence (i.e., labeling every batch), causing annotation cost to scale linearly with deployment length. As a result, the complementary problem of deciding \emph{when} to request supervision under a global budget remains largely unexplored. 
Our work addresses this gap by explicitly reasoning about supervision allocation over time.
\section{Conclusion}
\label{conclusion}

In this work, we study \emph{budgeted} active test-time adaptation, where supervision is available only for a fraction of test batches and must be allocated over time.
We introduce WISE-ATTA, which jointly decides \emph{when} to query labels via budget-paced batch selection and \emph{what} to label via drift-based single-sample querying.
Across synthetic corruptions and natural distribution shifts, WISE-ATTA achieves competitive or improved robustness compared to prior ATTA methods while using substantially fewer labels.
More broadly, our results underscore the importance of \emph{temporal} supervision allocation for stable and label-efficient test-time adaptation.

\section*{Acknowledgment}

This work was funded by the German Federal Ministry of Education and Research under the grant AIgenCY (16KIS2012) and SisWiss (16KIS2330). In addition, this work was funded by zukunft.niedersachsen, the joint science funding program of the Lower Saxony Ministry of Science and Culture and the Volkswagen Foundation and the Daimler and Benz Foundation under the grant Ladenburger Kolleg, Project KonCheck.

{
    \small
    \bibliographystyle{style/ieeenat_fullname}
    \bibliography{reference}
}

\clearpage
\appendix
\onecolumn

\clearpage
\appendix
\crefname{section}{Appendix}{Appendices}
\Crefname{section}{Appendix}{Appendices}

\section*{Appendix Overview}

This appendix provides additional analyses, deployment studies, and
implementation details supporting the main paper. It is organized into three
parts.

\paragraph{Part I --- Method Analysis and Ablations.}
We first examine the design and behavior of WISE-ATTA. We clarify the distinct
roles of batch- and sample-level supervision
(\cref{app:batch_vs_sample}), compare alternative batch-utility functions
(\cref{sec:utility}), analyze the behavior of queried samples
(\cref{app:query_behavior}), study sensitivity to the batch-selection
hyperparameters (\cref{app:hyper_sensitivity}), and provide the complete
batch-selection results underlying the main-paper analysis
(\cref{app:batch_selection_results}).

\paragraph{Part II --- Practical Deployment Studies.}
We then examine WISE-ATTA under several deployment constraints: skipping
updates on non-selected batches (\cref{app:skip_updates}), varying online
batch sizes (\cref{app:batchsize}), and delayed label availability
(\cref{app:annotation_delay}).

\paragraph{Part III --- Implementation and Experimental Details.}
Finally, we provide the complete algorithm and hyperparameters
(\cref{app:implementation}), analyze which parameters should receive
supervised updates (\cref{app:layers}), and provide dataset, model, and
reproducibility details (\cref{app:experimental_details}).

\part*{Part I: Method Analysis and Ablations}

\section{Batch-Level vs.\ Sample-Level Supervision}
\label{app:batch_vs_sample}

WISE-ATTA makes two separate supervision decisions: \emph{whether} to select
the current batch for supervision, and \emph{which} sample to label once that
batch is selected. Since these decisions address different questions, they rely
on different signals.

\paragraph{Ideal batch.}
An ideal batch for supervision is not the one with the most uncertain
predictions, where supervised gradients are likely to be noisy, nor the one
where the model is already converged, where supervision is wasted. Instead, it
is a batch in which the model is already reasonably aligned with the current
target regime, so that a labeled update is likely to be reliable rather than
dominated by noise.
WISE-ATTA approximates this property using the number of low-entropy
predictions in the batch, which serves as a proxy for batch-level adaptation
stability: when many samples in the batch already receive relatively confident
predictions, supervision is more likely to reinforce ongoing adaptation than to
destabilize it.

\paragraph{Ideal sample.}
Within a selected batch, the goal is different. An ideal sample is neither the
most confident nor the most uncertain example; instead, it should still be
undergoing meaningful adaptation, so that a corrective label can have a
nontrivial effect, while not being so unstable that a single supervision signal
is unlikely to help.
WISE-ATTA approximates this property using prediction drift relative to an EMA
anchor, which singles out samples whose predictions are still changing in a
structured way that results in ongoing but not yet converged adaptation
dynamics.

Prediction entropy and prediction drift capture different aspects of model
behavior and therefore serve complementary roles. \Cref{fig:motivation_plots_2}
shows that WISE-ATTA tends to query samples in a moderate-entropy range rather
than the highest-entropy tail, consistent with this design.

\section{Impact of Batch Utility Functions}
\label{sec:utility}

We analyze how different definitions of batch utility affect adaptation under
constrained supervision. WISE-ATTA selects batches for annotation based on a
utility score, and we ablate several choices for measuring batch utility.
\Cref{tab:utility_ablation} compares entropy-based utilities (mean, top-$k$,
and low-$k$ entropy), drift-based utilities (mean and top-$k$ prediction
drift), and our default $\textsc{Count}(H<\tau)$.

\begin{table}[h]
\centering
\footnotesize
\caption{Effect of different batch utility functions (error \%, $\downarrow$).}
\label{tab:utility_ablation}
\setlength{\tabcolsep}{4pt}
\resizebox{0.50\linewidth}{!}{
\begin{tabular}{@{} l | cc | cc @{}}
\toprule
\textbf{Utility Function}
& \multicolumn{2}{c}{ImageNet-R}
& \multicolumn{2}{c}{ImageNet-K} \\
\cmidrule(lr){2-3} \cmidrule(lr){4-5}
& RN50-BN & ViT-B-16
& RN50-BN & ViT-B-16 \\
\midrule
Mean Entropy      & 52.82 & 44.81 & 65.54 & 58.31 \\
Top-$k$ Entropy   & 52.77 & 44.62 & 65.62 & 58.85 \\
Low-$k$ Entropy   & 52.88 & 44.26 & 65.46 & 58.69 \\
\midrule
Mean Drift        & 52.85 & 44.32 & 65.23 & 58.78 \\
Top-$k$ Drift     & 52.25 & 43.94 & 65.09 & 58.44 \\
\midrule
\rowcolor{gray!12}
Count($H < \tau$) & \textbf{52.26} & \textbf{43.38} & \textbf{64.60} & \textbf{57.95} \\
\bottomrule
\end{tabular}}
\end{table}

Overall, entropy-based utilities perform worse across settings, regardless of
whether we aggregate by mean, top-$k$, or low-$k$, suggesting that entropy
statistics alone are not a reliable proxy for batch utility. Among drift-based
variants, top-$k$ drift is consistently stronger than mean drift and all
entropy-based alternatives, likely because it better reflects whether a batch
contains a small number of high-value samples for supervised updates. Finally,
$\textsc{Count}(H<\tau)$ achieves the best performance overall, reducing error
by an average of $0.82$ points across datasets and backbones compared to using
\textsc{Mean Entropy}, suggesting that batches with more stable (low-entropy)
predictions tend to yield more reliable adaptation updates.

\section{Query Behavior}
\label{app:query_behavior}

\begin{figure}[h]
  \centering
  {\small\bfseries ResNet-50}
  \par\vspace{0.2em}

  \begin{subfigure}[b]{0.35\linewidth}
    \centering
    \includegraphics[width=\linewidth]{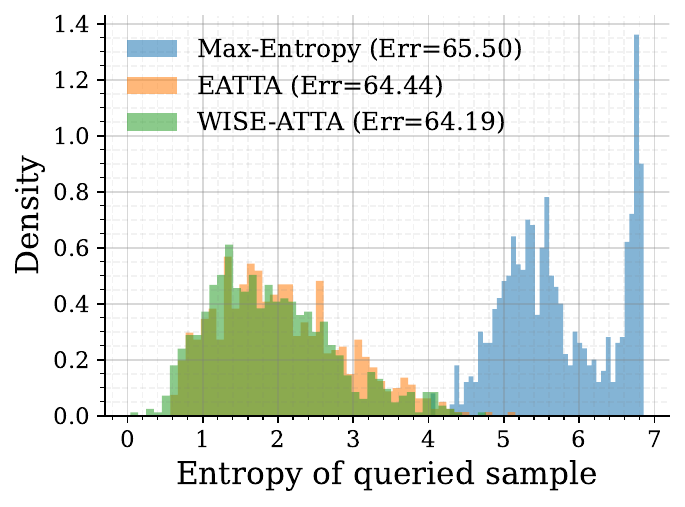}
    \caption{}
    \label{fig:motivation_rn_hist}
  \end{subfigure}\hspace{0.03\linewidth}
  \begin{subfigure}[b]{0.35\linewidth}
    \centering
    \includegraphics[width=\linewidth]{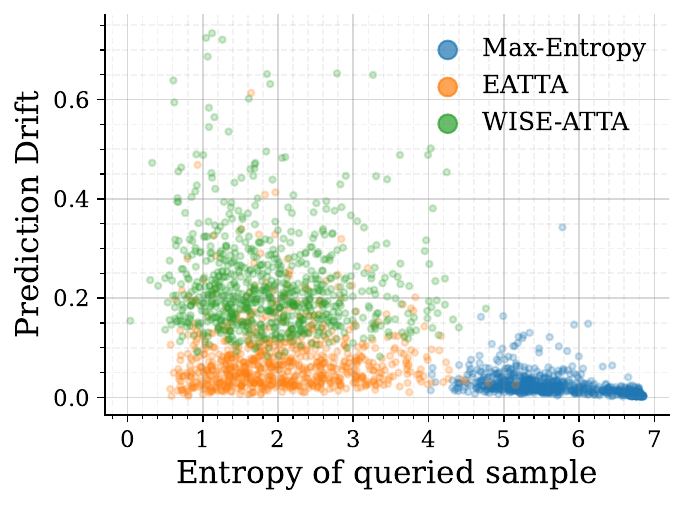}
    \caption{}
    \label{fig:motivation_rn_drift}
  \end{subfigure}

  \vspace{0.6em}

  {\small\bfseries ViT-B/16}
  \par\vspace{0.2em}

  \begin{subfigure}[b]{0.35\linewidth}
    \centering
    \includegraphics[width=\linewidth]{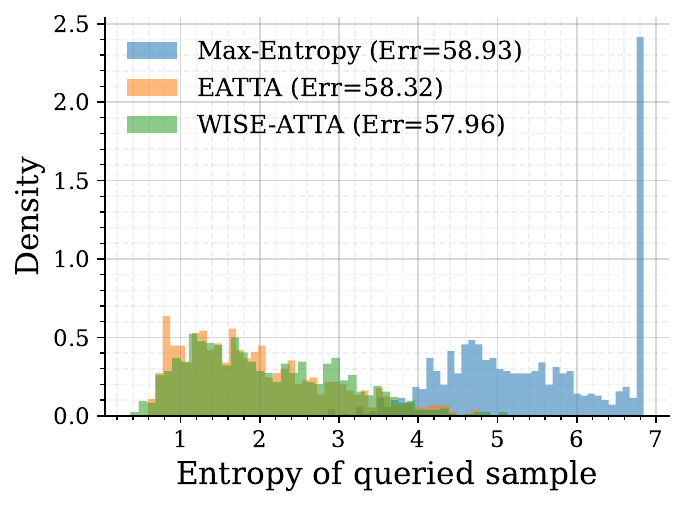}
    \caption{}
    \label{fig:motivation_vit_hist}
  \end{subfigure}\hspace{0.03\linewidth}
  \begin{subfigure}[b]{0.35\linewidth}
    \centering
    \includegraphics[width=\linewidth]{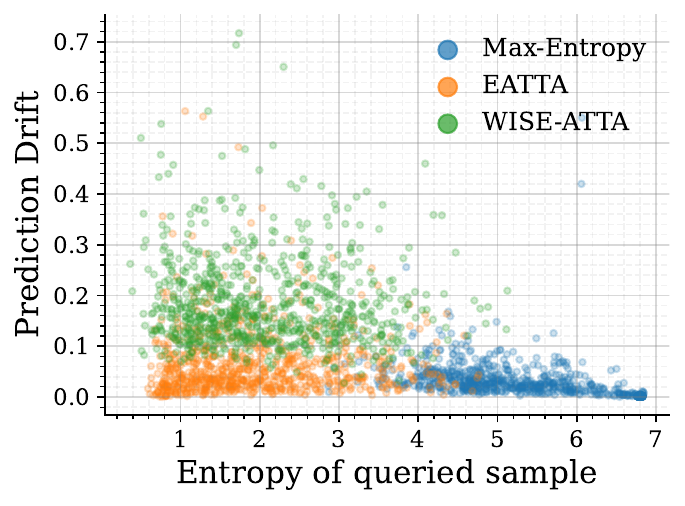}
    \caption{}
    \label{fig:motivation_vit_drift}
  \end{subfigure}

  \caption{Query behavior on ImageNet-R. Left: entropy distribution of queried samples;
right: prediction drift vs. entropy for samples queried by Max-Entropy, EATTA, and
WISE-ATTA. Prediction drift is the discrepancy between the current model and an exponential
moving average (EMA) anchor. Across RN50-BN (top) and ViT-B-16 (bottom), WISE-ATTA
avoids the highest-entropy extremes while favoring moderately uncertain samples with larger
drift.}
  \label{fig:motivation_plots_2}
\end{figure}

We further analyze the samples selected for annotation by different query
criteria on ImageNet-R. \Cref{fig:motivation_plots_2} shows (left) the entropy
distribution of queried samples and (right) the relationship between
queried-sample entropy and \emph{prediction drift}, defined as the discrepancy
between the current model and an exponential moving average (EMA) anchor.

\paragraph{Finding.}
\textsc{Max-Entropy} concentrates on extreme high-entropy samples, but these
samples exhibit near-zero drift, indicating high ambiguity with limited
directional signal for a single supervised update. \textsc{EATTA} shifts toward
lower-to-moderate entropy but largely remains in a low-drift band, suggesting
supervision is often spent where the model is already comparatively stable. In
contrast, \textsc{WISE-ATTA} selects a distinct regime of \emph{moderate
entropy with larger drift}, matching our goal of querying samples that are both
informative and learnable under a single-step update. This behavior is
consistent across RN50-BN and ViT-B-16 and coincides with the best performance.

\section{Sensitivity to Batch-Selection Hyperparameters}
\label{app:hyper_sensitivity}

We assess the sensitivity of WISE-ATTA to its batch-selection hyperparameters:
the history window size $W$, the warmup length $M$, and the slack parameter
$\delta$. Each hyperparameter is varied while the others are held fixed, and we
report error (\%) on ImageNet-R/K/A with RN50-BN and ViT-B-16 in
\cref{tab:hyper_sensitivity}.

\begin{table*}[h]
\centering
\caption{\textbf{Sensitivity to batch-selection hyperparameters} on ImageNet-R/K/A with RN50-BN and ViT-B-16 (FTTA error \%, $\downarrow$). The default value used throughout the paper is highlighted in gray. WISE-ATTA is robust across all three hyperparameters --- the spread in average error is at most $0.5$ points across each sweep.}
\label{tab:hyper_sensitivity}
\resizebox{0.85\linewidth}{!}{
\begin{tabular}{@{} l l | cc | cc | cc || c @{}}
\toprule
\multirow{2}{*}{\textbf{Setting}} & \multirow{2}{*}{\textbf{Value}}
& \multicolumn{2}{c}{ImageNet-R}
& \multicolumn{2}{c}{ImageNet-K}
& \multicolumn{2}{c}{ImageNet-A}
& \multirow{2}{*}{\textbf{Avg.}} \\
\cmidrule(lr){3-4} \cmidrule(lr){5-6} \cmidrule(lr){7-8}
&
& RN50-BN & ViT-B-16
& RN50-BN & ViT-B-16
& RN50-BN & ViT-B-16
& \\
\midrule
\multirow{2}{*}{Slack}
& w/o slack & 52.15 & 43.13 & 64.37 & 58.04 & 98.56 & 71.24 & 64.58 \\
& \cellcolor{gray!12}w/ slack
& \cellcolor{gray!12}52.03 & \cellcolor{gray!12}43.38
& \cellcolor{gray!12}64.29 & \cellcolor{gray!12}58.11
& \cellcolor{gray!12}98.41 & \cellcolor{gray!12}71.13
& \cellcolor{gray!12}\textbf{64.56} \\
\midrule
\multirow{6}{*}{History window $W$}
& $1$   & 52.59 & 44.40 & 65.01 & 58.50 & 97.89 & 71.84 & 65.04 \\
& $50$  & 52.06 & 44.07 & 65.16 & 58.10 & 98.32 & 70.81 & 64.75 \\
& $100$ & 52.42 & 43.70 & 64.91 & 58.35 & 98.39 & 70.57 & 64.72 \\
& $150$ & 52.16 & 43.54 & 65.13 & 58.07 & 98.39 & 70.57 & 64.64 \\
& \cellcolor{gray!12}$250$
& \cellcolor{gray!12}52.03 & \cellcolor{gray!12}43.38
& \cellcolor{gray!12}64.29 & \cellcolor{gray!12}58.11
& \cellcolor{gray!12}98.41 & \cellcolor{gray!12}71.13
& \cellcolor{gray!12}64.56 \\
& $300$ & 52.12 & 43.58 & 64.63 & 57.85 & 98.39 & 70.57 & \textbf{64.52} \\
\midrule
\multirow{4}{*}{Warmup length $M$}
& \cellcolor{gray!12}$1$
& \cellcolor{gray!12}52.03 & \cellcolor{gray!12}43.38
& \cellcolor{gray!12}64.29 & \cellcolor{gray!12}58.11
& \cellcolor{gray!12}98.41 & \cellcolor{gray!12}71.13
& \cellcolor{gray!12}\textbf{64.56} \\
& $5$  & 51.89 & 43.21 & 64.78 & 57.88 & 98.15 & 73.11 & 64.84 \\
& $10$ & 52.81 & 43.99 & 64.98 & 57.88 & 98.05 & 71.97 & 64.95 \\
& $15$ & 53.42 & 43.07 & 64.73 & 57.79 & 98.36 & 71.37 & 64.79 \\
\bottomrule
\end{tabular}
}
\end{table*}

\paragraph{Observation.}
Overall, WISE-ATTA is reasonably stable across a broad range of settings. The
average error varies only modestly across the tested values, suggesting that
the proposed batch-selection rule does not depend critically on fine-tuning
these hyperparameters.

\paragraph{Effect of slack $\boldsymbol{\delta}$.}
Adding slack yields a small but consistent improvement in the overall average
error ($64.58 \rightarrow 64.56$). This is consistent with the role of $\delta$
in \cref{alg:pubs}: the slack term avoids overly aggressive catch-up behavior
when label usage is only slightly behind the target rate. Without slack, the
controller is more reactive and may force supervision on batches whose utility
is not especially high; with slack, supervision can be reserved for more
informative batches.

\paragraph{Effect of history window $\boldsymbol{W}$.}
The method is fairly robust to the history window size. Very small windows,
especially $W=1$, perform worse, since the quantile threshold is then estimated
from too little history and becomes overly sensitive to short-term noise. As
$W$ increases, performance improves and stabilizes, with the best averages
obtained for larger windows such as $W=250$ and $W=300$. This suggests that
using a sufficiently long recent history provides a more reliable estimate of
relative batch utility while still adapting to changing stream conditions.

\paragraph{Effect of warmup length $\boldsymbol{M}$.}
Short warmup performs best, with $M=1$ giving the strongest average result
among the tested values. Increasing the warmup length gradually degrades
performance. This is expected because the warmup phase uses random budget
allocation before the utility threshold becomes active. A longer warmup
therefore delays the transition to utility-based selection and spends more of
the limited label budget without exploiting the batch-utility signal. In other
words, once even a small amount of history is available, it is better to begin
using the adaptive threshold rather than continue exploring randomly.

\section{Detailed Batch-Selection Results}
\label{app:batch_selection_results}

We provide the complete results underlying the batch-selection analysis. We report per-corruption results on
ImageNet-C under both CTTA and FTTA, as well as results on ImageNet-R/K/A
across RN50-BN and ViT-B-16.

\begin{table*}[h]
\centering
\caption{\textbf{ImageNet-C} error (\%) under CTTA (top) and FTTA (bottom) at a fixed annotation budget of $0.5$ labels per batch. We compare uniform, random, and utility-based batch selection under matched supervision budgets. Our method is highlighted in gray.}
\label{tab:imagenet-c-1}
\text{\tiny \textbf{CTTA Setting}}
\begin{threeparttable}
\resizebox{\linewidth}{!}{
\begin{tabular}{l l | ccc | cccc | cccc | cccc | c}
\toprule
\textbf{Sample Selection} & \textbf{Batch Selection}
& \multicolumn{3}{c}{Noise}
& \multicolumn{4}{c}{Blur}
& \multicolumn{4}{c}{Weather}
& \multicolumn{4}{c}{Digital}
& \textbf{Avg. Err.} \\
\cmidrule(lr){3-5}\cmidrule(lr){6-9}\cmidrule(lr){10-13}\cmidrule(lr){14-17}
&
& Gauss. & Shot & Impul.
& Defoc. & Glass & Motion & Zoom
& Snow & Frost & Fog & Brit.
& Contr. & Elastic & Pixel & JPEG
& \\
\midrule
\multirow{2}{*}{EATTA~\cite{wang2025effortless}}
& Uniform & 67.9 & 60.2 & 60.2 & 70.1 & 67.5 & 59.5 & 54.1 & 56.6 & 58.0 & 46.1 & 37.0 & 60.3 & 47.5 & 43.0 & 46.2 & 55.6 \\
& Random  & 66.7 & 59.6 & 59.6 & 69.4 & 66.5 & 59.1 & 53.7 & 55.7 & 58.3 & 46.1 & 37.1 & 60.8 & 47.3 & 43.3 & 46.2 & 55.3 \\
\midrule
\multirow{3}{*}{WISE-ATTA}
& Uniform & 66.8 & 59.1 & 59.4 & 69.1 & 66.0 & 58.5 & 53.3 & 55.3 & 58.2 & 46.3 & 37.1 & 57.6 & 47.2 & 42.8 & 46.1 & 54.9 \\
& Random  & 67.2 & 59.4 & 59.4 & 69.2 & 66.4 & 58.9 & 53.5 & 55.6 & 58.2 & 46.1 & 36.9 & 57.5 & 47.1 & 42.7 & 46.0 & 54.9 \\
& \cellcolor{gray!12}Budget-paced
& \cellcolor{gray!12}64.8 & \cellcolor{gray!12}58.8 & \cellcolor{gray!12}59.2 & \cellcolor{gray!12}68.1 & \cellcolor{gray!12}65.6 & \cellcolor{gray!12}58.1 & \cellcolor{gray!12}53.2 & \cellcolor{gray!12}54.4 & \cellcolor{gray!12}57.5 & \cellcolor{gray!12}45.3 & \cellcolor{gray!12}36.6 & \cellcolor{gray!12}56.4 & \cellcolor{gray!12}46.5 & \cellcolor{gray!12}42.3 & \cellcolor{gray!12}45.7 & \cellcolor{gray!12}\textbf{54.2} \\
\bottomrule
\end{tabular}
}
\end{threeparttable}

\text{\tiny \textbf{FTTA Setting}}
\begin{threeparttable}
\resizebox{\linewidth}{!}{
\begin{tabular}{l l | ccc | cccc | cccc | cccc | c}
\toprule
\textbf{Sample Selection} & \textbf{Batch Selection}
& \multicolumn{3}{c}{Noise}
& \multicolumn{4}{c}{Blur}
& \multicolumn{4}{c}{Weather}
& \multicolumn{4}{c}{Digital}
& \textbf{Avg. Err.} \\
\cmidrule(lr){3-5}\cmidrule(lr){6-9}\cmidrule(lr){10-13}\cmidrule(lr){14-17}
&
& Gauss. & Shot & Impul.
& Defoc. & Glass & Motion & Zoom
& Snow & Frost & Fog & Brit.
& Contr. & Elastic & Pixel & JPEG
& \\
\midrule
\multirow{2}{*}{EATTA~\cite{wang2025effortless}}
& Uniform & 67.9 & 64.0 & 65.8 & 74.3 & 70.0 & 53.5 & 48.5 & 49.6 & 55.6 & 41.1 & 32.2 & 72.3 & 43.7 & 40.0 & 45.7 & 54.9 \\
& Random  & 66.7 & 64.2 & 65.5 & 72.2 & 69.0 & 53.8 & 48.7 & 49.3 & 55.4 & 41.0 & 32.1 & 65.5 & 43.6 & 39.9 & 45.7 & 54.2 \\
\midrule
\multirow{3}{*}{WISE-ATTA}
& Uniform
& 66.8 & 64.0 & 65.9 & 70.1 & 69.4 & 54.3 & 48.8
& 49.3 & 55.3 & 41.0 & 32.2
& 60.2 & 44.0 & 40.0 & 45.6
& 53.8 \\
& Random
& 67.2 & 64.2 & 65.2 & 70.0 & 69.1 & 53.9 & 48.5
& 49.3 & 55.4 & 41.1 & 32.2
& 59.5 & 43.7 & 39.9 & 45.7
& 53.6 \\
& \gcell{Budget-paced}
& \gcell{64.8} & \gcell{62.4} & \gcell{63.5} & \gcell{67.5} & \gcell{67.3} & \gcell{51.9} & \gcell{47.8}
& \gcell{48.3} & \gcell{54.4} & \gcell{40.4} & \gcell{32.1}
& \gcell{57.5} & \gcell{42.8} & \gcell{39.4} & \gcell{44.9}
& \gcell{\textbf{52.3}} \\
\bottomrule
\end{tabular}
}
\end{threeparttable}
\end{table*}

\begin{table*}[h]
\caption{\textbf{Batch selection strategies under a fixed annotation budget.} FTTA error (\%) on ImageNet-R/K/A with RN50-BN and ViT-B-16.}
\label{tab:imagenet-variant-1}
\centering
\resizebox{0.88\linewidth}{!}{
\begin{tabular}{@{} l l | cc | cc | cc || cc}
\toprule
\textbf{Sample Selection} & \textbf{Batch Selection}
& \multicolumn{2}{c}{ImageNet-R}
& \multicolumn{2}{c}{ImageNet-K}
& \multicolumn{2}{c}{ImageNet-A}
& \multicolumn{2}{c}{\textbf{Avg. Error}} \\
\cmidrule(lr){3-4} \cmidrule(lr){5-6} \cmidrule(lr){7-8} \cmidrule(lr){9-10}
&
& RN50-BN & ViT-B-16
& RN50-BN & ViT-B-16
& RN50-BN & ViT-B-16
& RN50-BN & ViT-B-16 \\
\midrule
\multirow{2}{*}{EATTA~\cite{wang2025effortless}}
& Uniform
& 53.1 & 44.7 & 65.8 & 59.0 & 98.1 & 72.5 & 72.3 & 58.7 \\
& Random
& 53.1 & 44.8 & 65.6 & 58.8 & 98.6 & 72.4 & 72.4 & 58.7 \\
\midrule
\multirow{3}{*}{WISE-ATTA}
& Uniform
& 52.8 & 44.9 & 65.6 & 58.6 & 98.2 & 71.9 & 72.2 & 58.5 \\
& Random
& 52.6 & 44.6 & 65.4 & 58.3 & 98.4 & 71.8 & 72.2 & 58.2 \\
& \cellcolor{gray!12}Budget-paced
& \cellcolor{gray!12}52.2 & \cellcolor{gray!12}44.2
& \cellcolor{gray!12}65.3 & \cellcolor{gray!12}58.0
& \cellcolor{gray!12}98.2 & \cellcolor{gray!12}70.6
& \cellcolor{gray!12}\textbf{71.9} & \cellcolor{gray!12}\textbf{57.6} \\
\bottomrule
\end{tabular}
}
\end{table*}

\clearpage
\part*{Part II: Practical Deployment Studies}

\section{Skipping Updates on Non-selected Batches}
\label{app:skip_updates}

We ablate a low-resource variant of WISE-ATTA that performs \emph{no}
backward/update step on batches not selected by WISE-ATTA (i.e., no
unsupervised entropy-minimization update on non-selected batches).
\Cref{tab:skip_nonselected_updates} compares this setting (``without update'')
against the default configuration that still applies the unsupervised update
(``with update'') at a label ratio of $r{=}0.2$.

\begin{table}[h]
\centering
\caption{\textbf{Effect of skipping unsupervised updates on non-selected batches} at label ratio $r = 0.2$ (FTTA error \%, $\downarrow$). \emph{With update} applies the unsupervised loss on non-selected batches; \emph{Without update} skips the backward pass entirely. WISE-ATTA degrades the least when unsupervised updates are removed ($+0.1$ vs.\ $+0.2$ to $+0.4$ for the baselines), indicating that its gains do not rely on the unsupervised signal from skipped batches.}
\label{tab:skip_nonselected_updates}
\resizebox{0.75\linewidth}{!}{
\begin{tabular}{@{} l l | cc | cc || c c @{}}
\toprule
\multirow{2}{*}{\textbf{Setting}} & \multirow{2}{*}{\textbf{Method}}
& \multicolumn{2}{c}{ImageNet-R}
& \multicolumn{2}{c}{ImageNet-K}
& \multirow{2}{*}{\textbf{Avg.}}
& \multirow{2}{*}{$\Delta$ \textbf{Avg.}} \\
\cmidrule(lr){3-4} \cmidrule(lr){5-6}
&
& RN50-BN & ViT-B-16
& RN50-BN & ViT-B-16
& & \\
\midrule
\multirow{3}{*}{With update}
& Uniform   & 55.0 & 48.1 & 68.3 & 60.8 & 58.0 & -- \\
& Random    & 54.5 & 48.7 & 68.2 & 60.2 & 57.9 & -- \\
& \cellcolor{gray!12}WISE-ATTA
& \cellcolor{gray!12}\textbf{53.4} & \cellcolor{gray!12}\textbf{44.7}
& \cellcolor{gray!12}\textbf{66.2} & \cellcolor{gray!12}\textbf{58.7}
& \cellcolor{gray!12}\textbf{55.7} & \cellcolor{gray!12}-- \\
\midrule
\multirow{3}{*}{Without update}
& Uniform   & 55.0 & 48.2 & 68.6 & 61.0 & 58.2 & $+0.2$ \\
& Random    & 55.9 & 48.1 & 68.4 & 60.8 & 58.3 & $+0.4$ \\
& \cellcolor{gray!12}WISE-ATTA
& \cellcolor{gray!12}\textbf{53.8} & \cellcolor{gray!12}\textbf{45.0}
& \cellcolor{gray!12}\textbf{65.7} & \cellcolor{gray!12}\textbf{58.8}
& \cellcolor{gray!12}\textbf{55.8} & \cellcolor{gray!12}$+0.1$ \\
\bottomrule
\end{tabular}
}
\end{table}

Overall, removing updates on non-selected batches has negligible impact on
performance. For WISE-ATTA, the average error changes only from
$55.72 \rightarrow 55.82$ ($+0.10$), while Uniform and Random change by
similarly small amounts ($+0.18$ and $+0.41$, respectively). This indicates
that most of the gains come from spending computation and supervision on
high-utility batches, and that WISE-ATTA can be simplified for
resource-constrained deployment by updating only the selected batches.
Importantly, WISE-ATTA retains its advantage over Uniform/Random in this
lightweight mode, suggesting that budget-aware batch selection remains
effective even without background unsupervised updates.

\section{Batch Selection Under Different Online Batch Sizes}
\label{app:batchsize}

In practical deployments, the online batch size may vary due to latency and
memory constraints, which can affect both adaptation dynamics and batch utility
estimation. We therefore evaluate WISE-ATTA under different batch sizes on
ImageNet-R and ImageNet-K with RN50-BN and ViT-B-16 (\cref{tab:batch_size}).

\begin{table}[h]
\centering
\caption{\textbf{Batch-size ablation} on ImageNet-R/K with RN50-BN and ViT-B-16 (FTTA error \%, $\downarrow$). WISE-ATTA consistently outperforms uniform and random batch selection across all batch sizes.}
\label{tab:batch_size}
\resizebox{0.70\linewidth}{!}{
\begin{tabular}{@{} c l | cc | cc || cc @{}}
\toprule
\multirow{2}{*}{\textbf{Batch}} & \multirow{2}{*}{\textbf{Method}}
& \multicolumn{2}{c}{ImageNet-R}
& \multicolumn{2}{c}{ImageNet-K}
& \multicolumn{2}{c}{\textbf{Avg. Error}} \\
\cmidrule(lr){3-4} \cmidrule(lr){5-6} \cmidrule(lr){7-8}
&
& RN50-BN & ViT-B-16
& RN50-BN & ViT-B-16
& RN50-BN & ViT-B-16 \\
\midrule
\multirow{3}{*}{16}
& Uniform   & 56.5 & 41.8 & 56.5 & 41.8 & 56.5 & 41.8 \\
& Random    & 57.5 & 42.1 & 57.5 & 42.1 & 57.5 & 42.1 \\
& \cellcolor{gray!12}WISE-ATTA
& \cellcolor{gray!12}\textbf{55.1} & \cellcolor{gray!12}\textbf{41.5}
& \cellcolor{gray!12}\textbf{55.1} & \cellcolor{gray!12}\textbf{41.5}
& \cellcolor{gray!12}\textbf{55.1} & \cellcolor{gray!12}\textbf{41.5} \\
\midrule
\multirow{3}{*}{32}
& Uniform   & 53.0 & 43.2 & 65.1 & 57.5 & 59.0 & 50.4 \\
& Random    & 52.7 & 42.9 & 65.2 & 57.6 & 59.0 & 50.3 \\
& \cellcolor{gray!12}WISE-ATTA
& \cellcolor{gray!12}\textbf{52.1} & \cellcolor{gray!12}\textbf{41.9}
& \cellcolor{gray!12}\textbf{64.5} & \cellcolor{gray!12}\textbf{57.1}
& \cellcolor{gray!12}\textbf{58.3} & \cellcolor{gray!12}\textbf{49.5} \\
\midrule
\multirow{3}{*}{64}
& Uniform   & 52.8 & 44.4 & 65.6 & 58.6 & 59.2 & 51.5 \\
& Random    & 52.9 & 45.4 & 65.6 & 58.5 & 59.2 & 52.0 \\
& \cellcolor{gray!12}WISE-ATTA
& \cellcolor{gray!12}\textbf{52.3} & \cellcolor{gray!12}\textbf{43.4}
& \cellcolor{gray!12}\textbf{64.6} & \cellcolor{gray!12}\textbf{58.0}
& \cellcolor{gray!12}\textbf{58.4} & \cellcolor{gray!12}\textbf{50.7} \\
\midrule
\multirow{3}{*}{128}
& Uniform   & 54.0 & 47.6 & 67.2 & 59.8 & 60.6 & 53.7 \\
& Random    & 54.1 & 47.1 & 67.5 & 59.3 & 60.8 & 53.2 \\
& \cellcolor{gray!12}WISE-ATTA
& \cellcolor{gray!12}\textbf{53.1} & \cellcolor{gray!12}\textbf{45.7}
& \cellcolor{gray!12}\textbf{65.8} & \cellcolor{gray!12}\textbf{58.8}
& \cellcolor{gray!12}\textbf{59.4} & \cellcolor{gray!12}\textbf{52.2} \\
\bottomrule
\end{tabular}
}
\end{table}

Across all batch sizes, WISE-ATTA consistently outperforms \textsc{Uniform} and
\textsc{Random} batch selection, demonstrating that budget-paced selection
remains effective as the granularity of online updates changes. The
improvements are particularly evident for smaller batches. For example, at
batch size $16$, it achieves the lowest error across all settings (e.g.,
$55.08$ vs.\ $56.45/57.51$ on ImageNet-R (RN50-BN) and $41.53$ vs.\
$41.77/42.14$ on ImageNet-R (ViT-B-16)). Although the gap narrows at larger
batch sizes, it remains the best-performing strategy throughout (e.g., at batch
size $128$, $53.06$ vs.\ $53.98/54.07$ on ImageNet-R (RN50-BN)).

\section{Effect of Label Annotation Delay}
\label{app:annotation_delay}

The analyses thus far have assumed an idealization that is rarely true in
practice: that a queried label is returned to the learner instantaneously.
We now zoom out from this assumption and study a setting that, despite its
practical importance, has remained largely underexplored in active TTA:
\emph{label annotation delay}, where supervision arrives only after a fixed
number of subsequent test batches.
This better reflects realistic deployments, where human annotators and large
teacher-model incur queueing and inference latency.
To disentangle whether any observed degradation is specific to the front-loaded
budget allocation or general to ATTA, we compare three methods that change one
component at a time: \textsc{EATTA}, \textsc{WISE-ATTA + Uniform}, and full
\textsc{WISE-ATTA}. The first two share uniform batch allocation and differ only
in sample selection, while the last two share drift-based sample selection and
differ only in batch selection.
We
evaluate on ImageNet-R and ImageNet-K.

\begin{table*}[ht]
\centering
\caption{\textbf{Effect of annotation delay} at a fixed labeling rate of $r=0.5$. FTTA error (\%, $\downarrow$) on ImageNet-R and ImageNet-K with RN50-BN and ViT-B-16, as a function of the delay (in batches) between batch selection and label availability. WISE-ATTA is highlighted in gray.}
\label{tab:label_delay}
\resizebox{0.95\linewidth}{!}{
\begin{tabular}{@{} c | ccc | ccc || ccc | ccc @{}}
\toprule
\multirow{3}{*}{\textbf{Delay}}
& \multicolumn{6}{c||}{\textbf{ImageNet-R}}
& \multicolumn{6}{c}{\textbf{ImageNet-K}} \\
\cmidrule(lr){2-7} \cmidrule(lr){8-13}
& \multicolumn{3}{c|}{RN50-BN}
& \multicolumn{3}{c||}{ViT-B-16}
& \multicolumn{3}{c|}{RN50-BN}
& \multicolumn{3}{c}{ViT-B-16} \\
\cmidrule(lr){2-4} \cmidrule(lr){5-7} \cmidrule(lr){8-10} \cmidrule(lr){11-13}
& EATTA & WISE+Unif. & \cellcolor{gray!12}WISE-ATTA
& EATTA & WISE+Unif. & \cellcolor{gray!12}WISE-ATTA
& EATTA & WISE+Unif. & \cellcolor{gray!12}WISE-ATTA
& EATTA & WISE+Unif. & \cellcolor{gray!12}WISE-ATTA \\
\midrule
0   & 54.1 & 53.1 & \cellcolor{gray!12}\textbf{52.3} & 47.0 & 44.5 & \cellcolor{gray!12}\textbf{43.4} & 65.6 & 65.6 & \cellcolor{gray!12}\textbf{64.6} & 60.3 & 59.0 & \cellcolor{gray!12}\textbf{58.0} \\
50  & 54.2 & 54.1 & \cellcolor{gray!12}\textbf{52.9} & 46.5 & \textbf{46.3} & \cellcolor{gray!12}47.2 & 67.0 & 66.1 & \cellcolor{gray!12}\textbf{65.3} & 81.3 & 68.4 & \cellcolor{gray!12}\textbf{67.9} \\
100 & 54.2 & 54.5 & \cellcolor{gray!12}\textbf{54.0} & \textbf{54.6} & 56.2 & \cellcolor{gray!12}57.8 & 66.7 & 66.7 & \cellcolor{gray!12}\textbf{65.2} & 85.8 & 74.1 & \cellcolor{gray!12}\textbf{71.9} \\
150 & 54.7 & 54.8 & \cellcolor{gray!12}\textbf{54.6} & 70.1 & \textbf{59.8} & \cellcolor{gray!12}62.9 & 66.8 & 66.7 & \cellcolor{gray!12}\textbf{65.8} & \textbf{88.3} & 89.3 & \cellcolor{gray!12}92.2 \\
200 & 54.9 & 54.9 & \cellcolor{gray!12}\textbf{54.6} & 70.7 & \textbf{69.2} & \cellcolor{gray!12}70.5 & \textbf{67.0} & 67.1 & \cellcolor{gray!12}66.2 & \textbf{89.1} & 89.9 & \cellcolor{gray!12}93.0 \\
\bottomrule
\end{tabular}
}
\end{table*}

\paragraph{Results.}
\Cref{tab:label_delay} reveals two interesting findings.
First, all three methods degrade systematically as delay grows, confirming that
stale supervision is a general failure mode of ATTA rather than an artifact of
utility-driven scheduling.
Second, despite this shared degradation, full \textsc{WISE-ATTA} still achieves
the lowest error in $14$ of $20$ dataset model delay combinations, indicating
that budget-paced batch selection retains its advantage across most of the
delay regime.
Degradation is most severe on ViT-B-16/ImageNet-K, where error rises from
$58.0$ to $93.0$ as delay grows from $0$ to $200$.
This indicates that once the model has drifted far from the query-time regime,
stale labels can become actively harmful, and motivates delay-aware extensions
such as recency-weighted updates or latency-aware batch selection as an
important direction for future work.

\clearpage
\part*{Part III: Implementation and Experimental Details}

\section{Algorithm and Hyperparameters}
\label{app:implementation}

\begin{algorithm}[h]
\caption{Budget-Paced Utility Batch Selection (\textsc{WISE-ATTA})}
\label{alg:pubs}
\begin{algorithmic}[1]
\REQUIRE Stream $\{\mathcal{B}_t\}_{t\ge 1}$; target ratio $r\in[0,1]$; window $W$; warmup $M$; correction horizon $H_c$; slack $\delta$
\STATE $u_0 \leftarrow 0$,\; $\mathcal{H} \leftarrow [\ ]$ \COMMENT{$u_t$: labels used; $\mathcal{H}$: recent scores}
\FOR{$t = 1, 2, \dots$}
    \STATE $s_t \leftarrow \textsc{Utility}(\mathcal{B}_t)$ \COMMENT{batch utility, Eq.~\eqref{eq:utility}}
    \STATE $u_t^\star \leftarrow r\,t$ \COMMENT{target cumulative usage}
    \IF{$u_{t-1} + \delta < u_t^\star$}
        \STATE $a_t \leftarrow 1$ \COMMENT{rate-floor: catch up}
    \ELSIF{$|\mathcal{H}| < M$}
        \STATE $a_t \sim \mathrm{Bernoulli}(r)$ \COMMENT{warmup}
    \ELSE
        \STATE $d_t \leftarrow rt - u_{t-1}$;\; $\tilde{r}_t \leftarrow \mathrm{clip}(r + d_t/H_c, 0, 1)$ \COMMENT{debt, effective rate}
        \STATE $\tau_t \leftarrow \mathrm{Quantile}_{\,1-\tilde{r}_t}(\mathcal{H})$;\; $a_t \leftarrow \mathbb{I}[s_t \ge \tau_t]$ \COMMENT{select if high-utility}
    \ENDIF
    \STATE $u_t \leftarrow u_{t-1} + a_t$;\; append $s_t$ to $\mathcal{H}$ (drop oldest if $|\mathcal{H}| > W$)
\ENDFOR
\end{algorithmic}
\end{algorithm}

\paragraph{Hyperparameters.}
Unless stated otherwise, we use a batch size of $64$, following prior
work~\cite{wang2025effortless}; we study the sensitivity to batch size in
\cref{app:batchsize}.
Learning rates are set to $2.5\times10^{-4}$ for ImageNet-C, $10^{-3}$ for
ImageNet-R/K, and $5\times10^{-3}$ for ImageNet-A.
We use an EMA momentum of $\mu=0.9$ (\cref{pds-eq}), a history window of
$W=250$, warmup length $M=1$, correction horizon $H_c=50$, and slack
$\delta=1$ (\cref{alg:pubs}).
The combined loss in \cref{eq:total_loss} uses $\lambda_{\mathrm{sup}}=0.9$ and
$\lambda_{\mathrm{ent}}=0.1$ and is optimized with SGD.
We set the entropy threshold to $\tau_{\mathrm{ent}}=0.4\,\ln(C)$, where $C$ is
the number of classes.
All results are averaged over three random seeds and obtained on a server
equipped with a NVIDIA L40 GPU.

\section{Supervised Update: Which Parameters to Adapt?}
\label{app:layers}

\paragraph{Which parameters to update.}
In unsupervised test-time adaptation, prior work has shown that restricting
updates to normalization parameters is often sufficient and yields stable
behavior over long test
streams~\cite{wang2020tent,wang2025effortless,gui2024active}.
In the budgeted ATTA setting, however, supervision can be explicitly
corrective: a queried label provides reliable information about the decision
boundary. Restricting such supervision to normalization parameters alone can
thus limit its impact. Therefore, we adopt a hybrid strategy. In unlabeled
batches, we update only normalization parameters using entropy minimization.
When a labeled sample is available, we additionally allow a supervised update
of the classifier head.
We validate this below.

\begin{figure}[t]
  \centering
  \begin{subfigure}[b]{0.42\linewidth}
    \centering
    \includegraphics[width=\linewidth]{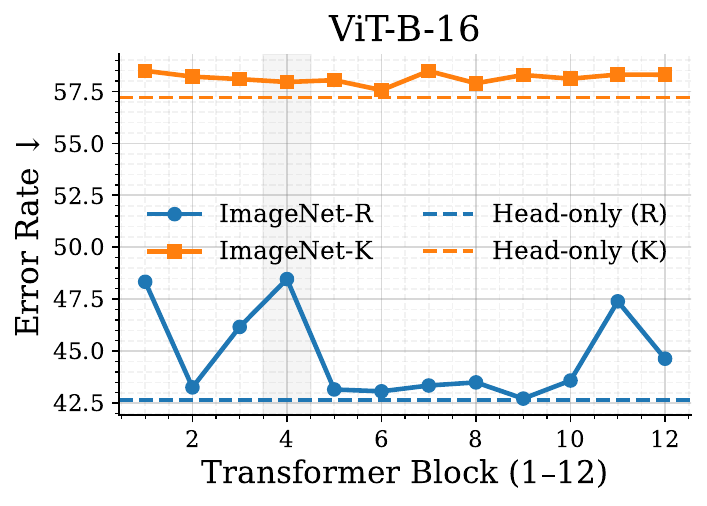}
    \caption{ViT-B-16}
    \label{fig:layer_ablation_vit}
  \end{subfigure}\hfill
  \begin{subfigure}[b]{0.42\linewidth}
    \centering
    \includegraphics[width=\linewidth]{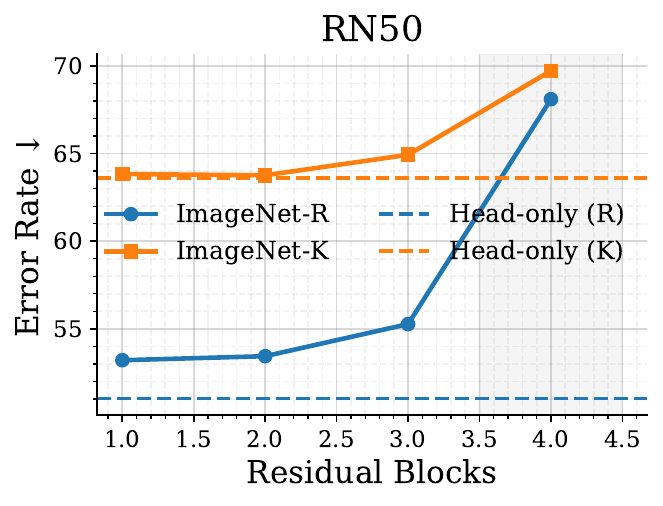}
    \caption{RN50-BN}
    \label{fig:layer_ablation_rn50}
  \end{subfigure}
  \caption{\textbf{Layer ablation for the supervised step in two-step
  adaptation.} Updating only the classifier head is most stable, while adapting
  intermediate/deeper layers often hurts under single-label supervision.}
  \label{fig:layer_ablation}
\end{figure}

\paragraph{Layer analysis for the supervised step.}
We investigate where supervised cross-entropy is most beneficial when only a
single labeled sample is available. After the first-step normalization
adaptation, we apply the supervised update to different model components and
report the resulting error (\cref{fig:layer_ablation}).

Across architectures and datasets, restricting supervision to the classifier
head yields the most stable performance. For ViT-B-16, updating intermediate
transformer blocks leads to high variance and is often worse than head-only
supervision. For RN50-BN, adapting deeper residual stages consistently degrades
accuracy, with error increasing sharply when supervision is applied to later
blocks. These results suggest that under limited supervision, pushing
supervised gradients into deeper layers can distort pretrained representations,
whereas head-only supervision improves robustness.

\section{Experimental Setup}
\label{app:experimental_details}

\subsection{Datasets and Models}
\label{app:datasets_models}

\paragraph{Datasets.}
We consider both synthetic corruptions (ImageNet-C) and natural distribution
shifts (ImageNet-R/K/A), which target complementary axes of robustness.
Low-level perturbations of the imaging pipeline (noise, blur, weather,
compression) and high-level shifts in rendition and source distribution, and
together span the corruption types on which prior ATTA work is evaluated.
For the synthetic corruptions, we consider
ImageNet-C~\cite{hendrycks2019benchmarking}, which consists of $15$ corruption
types at five severity levels; following prior
work~\cite{wang2025effortless,li2024exploring,niu2022efficient,niu2023towards,wang2020tent},
we report results at severity level $5$.
To natural corruptions, we use ImageNet-R~\cite{hendrycks2021many},
ImageNet-K~\cite{wang2019learning}, and ImageNet-A~\cite{hendrycks2021nae},
which contain renditions, sketches, and adversarially filtered images,
respectively. We construct the test streams by sequentially traversing the full
evaluation sets of each dataset, and report results on these complete streams
(\cref{app:data}).

\paragraph{Models.}
We consider two widely used vision models: ResNet-50 with Batch Normalization
(RN50-BN)~\cite{he2016deep,ioffe2015batch} and
ViT-B-16~\cite{dosovitskiy2020image}, both pretrained on
ImageNet-1K~\cite{5206848}.
We always start from the same pretrained checkpoint with no access to
source-domain data during adaptation.

\subsection{Reproducibility}
\label{app:repro}

 The main paper describes the
complete experimental setup, including datasets and evaluation protocols
(ImageNet-C/R/K/A under FTTA/CTTA), model architectures and pretrained
checkpoints (RN50-BN and ViT-B/16), optimization hyperparameters (batch size,
learning rates, SGD settings, and loss weights), and all method-specific
settings (e.g., $W$, $M$, $H$, $\delta$, $\tau_{\mathrm{ent}}$, and $\mu$). We
report results averaged over three random seeds ($0$, $41$, $58$). Code, configuration files, and evaluation scripts are available at \url{https://github.com/Muhammad-Huzaifaa/WISE-ATTA}.

\subsection{Dataset Details}
\label{app:data}

We evaluate our methods on both synthetic corruptions and natural distribution
shifts using four ImageNet-based benchmarks.

\paragraph{ImageNet-C.}
ImageNet-C~\cite{hendrycks2019benchmarking} applies $15$ common corruptions to
the ImageNet-1K validation set ($50{,}000$ images), grouped into Noise, Blur,
Weather, and Digital categories. Each corruption is provided at five severity
levels, yielding $750{,}000$ images per severity level and $3{,}750{,}000$ in
total. Following standard practice, we report results on severity level $5$.

\paragraph{ImageNet-R.}
ImageNet-R~\cite{hendrycks2021many} evaluates robustness to changes in
depiction style by collecting artistic renditions (e.g., cartoons, paintings,
sculptures) of ImageNet categories. It contains $30{,}000$ images spanning
$200$ classes.

\paragraph{ImageNet-K (Sketch).}
ImageNet-Sketch~\cite{wang2019learning} contains hand-drawn sketch
representations of ImageNet objects. It includes sketches for $1{,}000$
ImageNet classes with $50{,}000$ images in total; we refer to this benchmark as
ImageNet-K for consistency with our tables.

\paragraph{ImageNet-A.}
ImageNet-A~\cite{hendrycks2021nae} is a collection of natural adversarial
examples that remain recognizable to humans but are challenging for
ImageNet-trained models. It contains $7{,}500$ images from $200$ classes.

\end{document}